\documentclass[preprint,12pt,nonatbib]{elsarticle}

\usepackage{amssymb}
\usepackage{amsmath}
\usepackage{hyperref}
\usepackage{multirow}
\usepackage{subfig}
\usepackage{graphicx,url}
\usepackage{booktabs}

\journal{Information Sciences}

\begin{document}

\begin{frontmatter}



\title{Enhancing Classifier Evaluation: A Fairer Benchmarking Strategy Based on Ability and Robustness}


\author[UFPA,ITV]{Lucas Felipe Ferraro Cardoso\corref{cor1}}
\ead{lucas.cardoso@icen.ufpa.br}
\cortext[cor1]{Corresponding Author}

\author[UFPA,ITV]{Vitor Cirilo Araujo Santos}
\ead{vitor.cirilo.santos@itv.org}

\author[UFPA,ITV,IFPA]{José de Sousa Ribeiro Filho}
\ead{jose.ribeiro@ifpa.edu.br}

\author[UFPA]{Regiane Silva Kawasaki Francês}
\ead{kawasaki@ufpa.br}

\author[UFPE]{Ricardo Bastos Cavalcante Prudêncio}
\ead{rbcp@cin.ufpe.br}

\author[UFPA,ITV]{Ronnie Cley de Oliveira Alves}
\ead{ronnie.alves@itv.org}

\affiliation[UFPA]{organization={Federal University of Pará - UFPA},
            city={Belém},
            postcode={66075-11}, 
            state={PA},
            country={Brazil}}

\affiliation[ITV]{organization={Vale Technological Institute - ITV},
            city={Belém},
            postcode={66055-09}, 
            state={PA},
            country={Brazil}}

\affiliation[IFPA]{organization={Federal Institute of Pará - IFPA},
            city={Belém},
            postcode={67125-000}, 
            state={PA},
            country={Brazil}}

\affiliation[UFPE]{organization={Federal University of Pernambuco - UFPE},
            city={Recife},
            postcode={50670-901}, 
            state={PE},
            country={Brazil}}

\begin{abstract}
Benchmarking is a fundamental practice in machine learning (ML) for comparing the performance of classification algorithms. However, traditional evaluation methods often overlook a critical aspect: the joint consideration of dataset complexity and an algorithm’s ability to generalize. Without this dual perspective, assessments may favor models that perform well on easy instances while failing to capture their true robustness. To address this limitation, this study introduces a novel evaluation methodology that combines Item Response Theory (IRT) with the Glicko-2 rating system, originally developed to measure player strength in competitive games. IRT assesses classifier ability based on performance over difficult instances, while Glicko-2 updates performance metrics—such as rating, deviation, and volatility—via simulated tournaments between classifiers. This combined approach provides a fairer and more nuanced measure of algorithm capability. A case study using the OpenML-CC18 benchmark showed that only 15\% of the datasets are truly challenging and that a reduced subset with 50\% of the original datasets offers comparable evaluation power. Among the algorithms tested, Random Forest achieved the highest ability score. The results highlight the importance of improving benchmark design by focusing on dataset quality and adopting evaluation strategies that reflect both difficulty and classifier proficiency.
\end{abstract}




\end{frontmatter}



\section{Introduction}
\label{sec1}

Machine Learning (ML) has been growing very fast in recent years, due to the large number of applications that adopt ML models to support relevant tasks in different domains. The types of learning algorithms range from unsupervised to supervised \cite{monard2003conceitos}. In this work, the focus is on supervised learning algorithms, more precisely on classification algorithms, which are commonly adopted for pattern recognition tasks in several applications \cite{domingos2012few}. In literature, there is a wide range of supervised algorithms, which adopt different learning strategies to induce models from data \cite{monard2003conceitos}. Additionally, datasets can have particular characteristics, in such a way that it is not always simple to choose the best algorithm to be adopted for a given dataset. In this context, it is usual to empirically evaluate algorithms in order to identify the most promising ones for the datasets at hand. In fact, empirically evaluating ML algorithms is crucial to understand the advantages and limitations of the available techniques. 

Two important issues can be pointed out for empirically evaluating algorithms: (1) the datasets adopted in experiments; and (2) the methodology of experiments itself. Concerning the first issue, a common research practice in ML is to evaluate algorithms using benchmark datasets from online repositories, like the UCI respository \cite{Dua:2019}. Following the growth of the area, in recent years different online dataset repositories have emerged, such as OpenML \cite{vanschoren2014openml}. Specifically in OpenML, researchers can share datasets and experimental results, such as the performance of a classifier against a dataset. In addition, OpenML has several reference benchmarks, such as OpenML Curated Classification 2018 (OpenML-CC18), a benchmark for classification tasks that has 72 datasets curated and standardized by the platform \cite{bischl2017openml}.

Regarding the methodology of experiments, ML models can be trained and tested by applying a specific  execution procedure (e.g., cross-validation) on each dataset and then evaluated by using evaluation metrics of interest (e.g., accuracy). This strategy, however, does not allow for an in-depth analysis of the real ability of the model. Some datasets from a benchmark may be favoring an algorithm, giving the false impression that the classifier is, in fact, the best in relation to the others \cite{martinez2016making}. The complexity of the dataset should be taken into account during the process of evaluating the performance of an algorithm. Also only the use of the classical evaluation metrics can not guarantee that the evaluation result is completely reliable. 
In fact, when using an aggregate measure like accuracy, classifier performance is simply averaged across instances in a test 
 set without considering that some instances may be harder than others to classify. So a classifier may provide good predictions only for the easy instances, while miserably failing on the most difficult ones. Therefore, it is important that other instance-wise metrics are applied to result in a more robust performance assessment, as aggregated metrics can easily give false impressions about capabilities when a benchmark is not well constructed \cite{burnell2023rethink}.

 But between data and classifiers, which is more important? According to \cite{domingos2012few}, even a simpler classifier can beat the best classifier if the first classifier has much more training data than the second. At the same time, ``data alone is not enough'', i.e., the learning algorithms adopted in a dataset are crucial and make all the difference in the final learning results. In this way, data and models are two sides of the same coin of ML experimentation. So, an important question to address is how to evaluate classifiers, considering the dataset's importance as well. In previous works \cite{prudencio2015analysis,martinez2016making,martinez2019item,song2021efficient}, Item Response Theory (IRT) was adopted to answer this question and to provide a robust approach that allows evaluating both datasets and classifiers. IRT has been widely adopted in psychometric tests to measure an individual's ability to correctly answer a set of items (e.g., questions) by
taking into account the difficulty of the items. High ability values are assigned for individuals that correctly solve the most difficult items, while maintaining a consistent performance in the easy items as well. By considering classifiers as individuals and test instances as items, it is then possible to apply the concepts of IRT in the ML field. Thus, when applying IRT in ML, it is possible to simultaneously evaluate datasets and algorithms already considering the complexity of the dataset when measuring the classifier performance.

Despite the advantages of applying IRT in ML, evaluating benchmarks and algorithms with IRT is still a not straightforward task. In the current work, we propose the combination of IRT with rating systems \cite{samothrakis2014predicting}, in order to summarize the IRT results across datasets in a benchmark. Rating systems are commonly used to assess the ``strength'' of an individual in a competition (e.g., chess), through a series of tournaments. Rating systems are adopted to measure how proficient an individual is in a given activity. In this sense, such systems and IRT have some concepts in common which are explored in our work. More specifically, in our work, the Glicko-2 \cite{glickman2012example} rating system was adopted in order to create a ranking of models to summarize the results obtained by IRT on multiple datasets. In our proposal, each dataset is seen as a tournament in which the classifiers compete to each other. Initially, for each dataset, IRT is appled to measure the ability of each classifier. Then, each pair of classifiers compete to each other and a winner score is assigned to the classifier with higher ability for that dataset. The recorded scores are used by the Glicko-2 system to update the rating value, the rating deviation and the volatility of each classifier. After all datasets in the benchmark are considered in the tournaments, the final rating values are returned to produce a raking of classifiers.

In order to verify the viability of the proposed methodology, we performed a case study using the OpenML-CC18 benchmark. Preliminary results were obtained by \cite{cardoso2020decoding}, where a set of 60 datasets from the OpenML-CC18 benchmark was considered. In the current work, we extended the case study by addressing some questions like: \textit{``Would it be possible to use the IRT estimators to choose the best benchmark subset within OpenML-CC18?''}; \textit{``Are there datasets within a benchmark that might not be really good for evaluating learning algorithms?''} For this, IRT is used to create subsets of OpenML-CC18 datasets, then the combination of IRT and Glicko-2 is applied to generate the classifier rankings. Next, each ranking is analyzed, considering the expected performance of the classifiers to evaluate and then choosing a possible subset that is more efficient than the original one. Furthermore, the relationship between data and models is studied by analyzing whether the IRT estimators are related to the intrinsic characteristics of the datasets themselves.

The main contributions of this work are summarized as follows:

\begin{itemize}
    \item The proposal of a new methodology to simultaneously evaluate the performance of algorithms and the difficulty of datasets, based on the combination of IRT and Glicko-2;
    \item Application of the proposed methodology to analyze existing problems in a known benchmark in OpenML;
    \item All implementations in the case study are provided in a single tool (decodIRT), developed to automate the process of evaluating datasets and algorithms via IRT.
    \item Although decodIRT uses OpenML datasets for practicality, it is also possible to use it with local datasets that the user has.
  
\end{itemize}

The rest of this work is organized as follows: Section 2 contextualizes the main issues covered in this work, more precisely about classical performance metrics, OpenML, Item Response Theory and the Glicko-2 system. Section 3 presents the related work and compares it with the present work. Section 4 presents the methodology used, explains how IRT and the Glicko-2 system were used. Section 5 discusses the results obtained. Section 6 presents the final considerations of the work.


\section{Background}

\subsection{Classifier Evaluation and Benchmarks}

 Experiments in ML usually rely on robust benchmarks of datasets for algorithm training and testing. In fact, the creation of appropriate benchmarks are key part of the research in ML. They are important pieces for the standardization of studies in the area, enabling the community to follow the progress over time, identifying which problems are still a challenge and which algorithms are the best ones for certain applications. The lack of standardized benchmarks results in many studies using their own sets of pre-processed datasets in their own way. This condition makes it difficult to compare and reproduce the results obtained by these studies \cite{bischl2017openml}.


In ML it is not enough just to train an algorithm, generate a model and start using it. It is very important to know if the model that was generated was really able to learn to classify correctly. For this, one can apply performance evaluation metrics commonly adopted in literature. There are different performance metrics and each one can be more interesting than the other depending on the aspect you want to evaluate \cite{kubat2017introduction}.

Accuracy and error rate are one of the most used classic metrics. However, the result of a single performance metric can be misleading and not correctly reflect the true capability of a classifier \cite{kubat2017introduction}. In \cite{ferri2009experimental} the authors experimentally analyzed the behavior of a total of 18 performance metrics. In the work, it is reinforced that the different performance metrics can generate different evaluations about the model's ability depending on the situation, that is, it depends on the data set used. For example, in situations where there is an imbalance of classes or the dataset has few instances, a given metric may be preferable over the others. Thus, it is important to choose one or more specific metrics that are suitable to evaluate the model, taking into account the inner data complexities of the experiment.

\subsection{OpenML}

OpenML is a repository that works as a collaborative environment, where ML researchers can automatically share detailed data and organize it to work more efficiently and collaborate on a global scale \cite{vanschoren2014openml}. It also allows ML tasks to be executed with the repository datasets using the preference algorithm and then share the results obtained within the platform, minimizing the double effort. In addition, OpenML also makes it possible for new datasets to be made available by users, challenging the community to run algorithms on the dataset using specific parameters to solve a given ML task (e.g., classification) \cite{vanschoren2014openml}. 

In this context, OpenML also has the advantage of providing several reference benchmarks, such as the OpenMLCC-18 \footnote{Link to access OpenML-CC18: \url{https://www.openml.org/s/99}}. The OpenML-CC18 is a classification benchmark composed of 72 existing OpenML datasets from mid-2018 and which aims to address a series of requirements to create a complete reference set. In addition, it includes several datasets frequently used in benchmarks published in recent years \cite{bischl2017openml}.

The properties used to filter the datasets are: (a) Number of instances between 500 and 100,000; (b) Number of features up to 5000; (c) At least two classes targeted, where no class has less than 20 instances in total; (d) The proportion between minority and majority classes must be above 0.05; (e) Datasets cannot have been artificially generated; (f) Datasets must allow for randomization through a 10-field cross-validation; (g) No dataset can be a subset of another larger dataset; (h) All datasets must have some source or reference available; (i) No dataset should be perfectly classifiable by a single feature; (j) No dataset should allow a decision tree to achieve 100\% accuracy in a 10-field cross-validation task; (k) Datasets cannot have more than 5000 features after a \textit{one-hot-encoding} process on categorical features; (l) The datsets cannot have been created by binarizing regression or multiclass tasks; (m) No dataset can be sparse \cite{bischl2017openml}.

Therefore, it is understood that OpenML has a lot to contribute to research in the field of machine learning. In the previous work \cite{cardoso2020decoding} an initial analysis of OpenML-CC18 was performed using IRT, which allowed the generation of new relevant metadata about the complexity and quality of the benchmark, such as the difficulty and discriminative power of the data. In this present work, we seek to deepen this analysis by looking for a subset of datasets within OpenML-CC18 that is as good or perhaps better than the original. Using IRT to find a more efficient benchmark subset that maintains the characteristics of the original.

\subsection{Item Response Theory}
\label{irt_section}

To assess the performance of individuals in a test, traditionally, the total number of correct answers is used to rank the individuals evaluated. Despite being common, this approach has limitations to assess the actual ability of an individual. On the other hand, IRT allows the assessment of latent characteristics of an individual that cannot be directly observed and aims to present the relationship between the probability of an individual correctly responding to an item and their latent traits, that is, their ability in the assessed knowledge area. One of the main characteristics of the IRT is to have the items as central elements and not the test as a whole. The performance of an individual is evaluated based on their ability to hit certain items of a test and not how many items they hit \cite{baker2001basics}.

The IRT is a set of mathematical models that seek to represent the probability of an individual to correctly answer an item based on the item parameters and the respondent's ability, where the greater the individual's ability, the higher the chance of correctly answering an item \cite{baker2001basics}.

Logistic models for dichotomous items\footnote{It is only considered whether the answer to the item is right or wrong.} are the most used in literature. Among them, there are basically three types of models, which differ by the number of item parameters used. These are known as 1, 2 and 3 parameter logistic models. The 3-parameter logistic model, called 3PL, is the most complete among the three, where the probability of an individual $j$ correctly answering an item $i$ given their ability is defined by the Equation \ref{eq:3pl}.

\begin{equation} \label{eq:3pl}
    P(U_{ij} = 1\vert\theta_{j}) = c_{i} + (1 - c_{i})\frac{1}{1+ e^{-a_{i}(\theta_{j}-b_{i})}}
\end{equation}

\noindent Where:

\begin{itemize}
\setlength\itemsep{.25cm}
  
  \item $U_ {ij}$ is the dichotomous response that can take the values 1 or 0, being 1 when the individual \textit{j} hits the item \textit{i} and 0 when he misses;
  
  \item $\theta_{j}$ is the ability of the individual \textit{j};
  
  
  \item $b_{i}$ is the item's difficulty parameter and indicates the location of the logistic curve;
  
  \item $a_{i}$ is the item's discrimination parameter, i.e., how much the item \textit{i} differentiates between good and bad respondents. This parameter indicates the slope of the logistic curve. The higher its value, the more discriminating the item is;
  
  \item$c_{i}$ is the guessing parameter, representing the probability of a casual hit. It is the probability that a respondent with low ability hits the item.

\end{itemize}

\begin{figure}[ht]
\centering
\includegraphics[width=.8\textwidth]{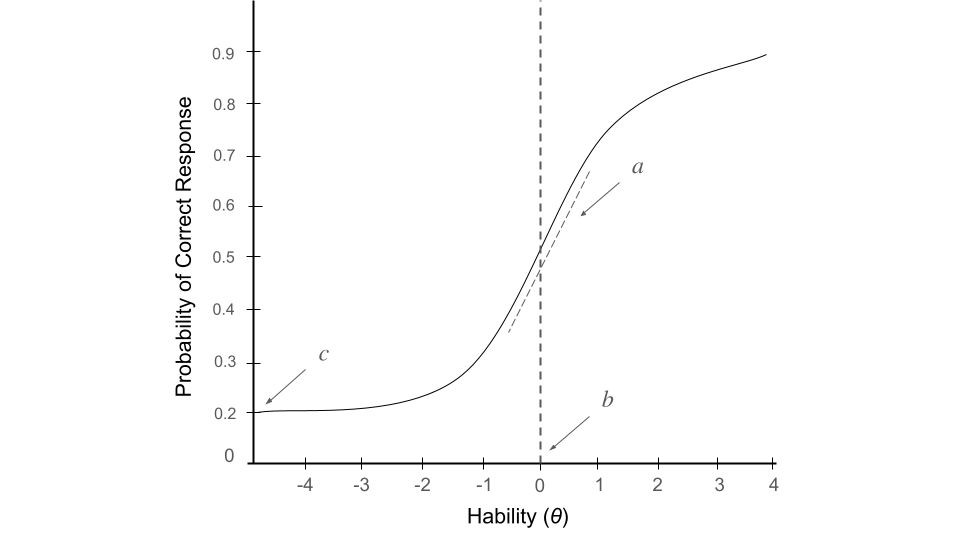}
\caption{Example of Item Characteristic Curve.}
\label{fig:icc_exemplo}
\end{figure}

The relationship between item parameters can be better understood by observing the Item Characteristic Curve (ICC), represented in Figure \ref{fig:icc_exemplo}. The ICC aims to demonstrate the variation in the probability of a respondent's correct answer to an item, considering the item parameters and the variation in ability. The Guessing parameter $c_{i}$ represents the minimum correct answer probability that a respondent will have on an item. The difficulty parameter $b_{i}$ represents the midpoint of the curve where the probability of a correct answer will be equal to 0.5, i.e. where the value of $\theta_{j}$ is equal to the difficulty value $b_{i} $. The discrimination parameter $a_{i}$ represents the slope of the curve, so that the greater the discrimination, the steeper the curve.

Typically $a_i >0$, in such a way that the probability of success is greater for individuals with higher ability values. Negative discrimination are not expected in a well designed test and can alert about a possible inconsistency in the item. So in practice items with negative discrimination are discarded.

The other two logistic models can be obtained by simplifying the 3PL. For 2PL the guessing parameter is removed, i.e., $c_{i}=0$. The guessing parameter also represents the lowest possible correct answer probability, so removing it eliminates the possibility that a low-ability respondent might respond correctly by luck. For 1PL the discrimination parameter is also removed, assuming that $a_{i}=1$. The discrimination parameter concerns the quality of the item itself, where negative discrimination can alert about a possible inconsistency, so when removing it, useful information about the item's integrity can be lost.

To estimate the parameters of the items and the ability of the respondents, the set of responses from all individuals is used for all the items that will be evaluated. For this, simple methods such as Maximum Likelihood Estimation (MLE) are commonly used, whether in cases where only the ability are known, only the item parameters are known or when none of them are known \cite{martinez2019item}.


IRT also has the concept of True-Score \cite{lord1984comparison}, which is the sum of the estimated correct probabilities across the items in the test. The True-Score is then used to set a final score that summarizes the respondent's performance. 

\begin{equation} \label{eq:true_score}
    TS_{j} = \sum_{i=1}^{N} P(U_{ij} = 1\vert\theta_{j})
\end{equation}

The True-Score is defined by the Equation \ref{eq:true_score}, where $TS_{j}$ is the True-Score calculated for a given respondent with ability level $\theta_{j}$ and $P(U_{ij} = 1\vert\theta_{j})$ is the probability of a correct answer calculated for each item $i$, depending on the logistic model used. Based on the above, it is understood that IRT can be an adequate approach to assess the real ability of classifiers and the complexity of datasets \cite{baker2001basics}.

\subsubsection{IRT True-Score Example}
\label{true_score_sub}

This section aims to exemplify how IRT estimators are used for evaluation. To this end, an example will be presented demonstrating how IRT is used to evaluate both respondents and the items that make up the test. Table \ref{tab:resp} presents a hypothetical set of 5 respondents who answered a test with 5 items. In the response table, a 1 is recorded for a correct answer and a 0 for an error.

\begin{table}[h]
\begin{center}
\caption{Example of a set of respondents.}\label{tab:resp}%
\begin{tabular}{@{}ccccccc@{}}
\toprule
& Item 1 & Item 2 & Item 3 & Item 4 & Item 5 & Total \\
\midrule
Individual 1 & 1      & 0      & 0      & 1      & 1      & 3     \\
Individual 2 & 0      & 0      & 1      & 1      & 0      & 2     \\
Individual 3 & 1      & 1      & 0      & 1      & 0      & 3     \\
Individual 4 & 0      & 0      & 1      & 1      & 1      & 3     \\
Individual 5 & 1      & 0      & 1      & 0      & 0      & 2         \\
\bottomrule
\end{tabular}
\end{center}
\end{table}

Following the classic evaluation methodology, where the total number of correct answers obtained by the respondent is only added, it is noted that the case presented has 3 respondents with equal scores. Therefore, it would not be possible to identify which is the best among the 3 without individually evaluating each item answered correctly or incorrectly. In this case, it would be possible due to the small number of items, but imagining an ML evaluation in which more than one model achieved the same score, there will probably be dozens or hundreds of instances in the test dataset to be evaluated individually and then indicate which model is most suitable.

\begin{table}[h]
\begin{center}
\caption{Example of a set of items.}\label{tab:items}%
\begin{tabular}{@{}cccc@{}}
\toprule
& Discrimination & Difficulty & Guessing \\
\midrule
Item 1 & 1.199         & -0.899      & 0.242       \\
Item 2 & 1.319         & 0.255       & 0.262       \\
Item 3 & 0.760         & -1.054      & 0.241       \\
Item 4 & 1.462         & -0.809      & 0.274       \\
Item 5 & 1.552         & -0.156      & 0.275         \\
\bottomrule
\end{tabular}
\end{center}
\end{table}

However, when applying IRT it is possible to explore individual items in a much more assertive way, by using the values of the item parameters. The Table \ref{tab:items} presents the Discrimination, Difficulty and Guessing values of each item in the example presented. And using only the Difficulty parameter it is already possible to understand the performance of individuals, it is noted that 4 items have negative difficulty values, this means that the test itself is mostly composed of items considered easy. Given this, one can imagine that the individual who gets the most difficult item right must be the best, but this is not what happens when we look at the final True-Score score in Table \ref{tab:true}.

\begin{table}[h]
\begin{center}
\caption{Example of True-Score.}\label{tab:true}%
\begin{tabular}{@{}cc@{}}
\toprule
& True-Score \\
\midrule
Individual 1 & 3.457      \\
Individual 2 & 2.214      \\
Individual 3 & 3.169      \\
Individual 4 & 3.063      \\
Individual 5 & 2.206         \\
\bottomrule
\end{tabular}
\end{center}
\end{table}

In Table \ref{tab:true} it is individual 1 who has the highest True-Score score, who in turn incorrectly answered item 2 considered the most difficult, while individual 3 who answered the most difficult item correctly has the second highest True-Score score. So, what explains this situation? This happens because IRT works based on the Principle of Consistent Order, which values consistency in individuals' responses, i.e., if a respondent gets an item of a certain difficulty right, then he or she must be able to get any item with a lower difficulty right. In a more practical example, it means that if you are able to jump to a certain height, you can then jump to any height lower than that limit. In this example, Individual 3's score is penalized because he correctly answers the most difficult item and incorrectly answers the easiest one, thus this situation violates the Principle of Consistent Order \cite{baker2001basics}.

As presented, the use of IRT is capable of expanding the way of evaluating a respondent using a test. This ability can be interesting to explore new paths in evaluating ML models and datasets, since commonly used metrics do not evaluate models at the instance level.

\subsection{Glicko-2 System}
\label{glicko_section}

Although IRT already has the True-Score calculation as its own metric to generate a final score, one of its objectives is to compare several classifiers on different dataset benchmarks. Performing so many comparisons at the instance level is not a trivial task, therefore it is understood that it is necessary to apply a more robust evaluation method together with IRT in order to properly explore the concept of classifiers' ability.

Rating systems are usually used in competitions to measure the ``strength'' of competitors, where each individual will have their own rating value and after a match this value is updated depending on the result (win, draw or defeat). Among the existing rating systems, Glicko-2 is the update of the Glicko system developed by Mark E. Glickman \cite{glickman2012example} to measure the strength of chess players. The Glicko system was developed in order to improve the \textit{Elo} system \cite{elo1978rating} taking into account the players' activity period to ensure greater reliability to the rating value \cite{samothrakis2014predicting}.

In the Glicko-2 system, each individual has three variables used to measure the statistical strength, they are: the rating value R, the rating deviation (RD) and the volatility ($\sigma$). Despite being very approximate, it cannot be said that the rating value perfectly measures an individual's ability, as it is understood that this value may suffer some variation. For this, the Glicko system has the RD, which allows calculating a 95\% reliable range of rating variation, using the formula: $[R-2RD, R+2RD]$. This means that there is a 95\% chance that the individual's actual strength is within the calculated range. Therefore, the smaller the RD value, the higher the rating precision \cite{glickman2012example,samothrakis2014predicting}.

To measure how much fluctuation the rating is within its RD range, Glicko uses volatility. Thus, the higher the volatility value, the greater the chances of the rating having large fluctuations within its range, and the lower the volatility, the more reliable the rating is. For example, in a dispute between individuals with low volatility values, based on their ratings it is possible to state more precisely who is the strongest \cite{samothrakis2014predicting,vevcek2014chess}.

The Glicko-2 system uses the concept of rating period to estimate rating values, which consist of a sequence of matches played by the individual. At the end of this sequence, the Glicko system updates the player's parameters using the opponents' rating and RD along with the results of each game (e.g., 1 point for victory and 0 for defeat). If the individual is being evaluated for the first time, the Glicko system uses standardized initial values, being: 1500 for rating, 350 for RD and 0.06 for volatility \cite{glickman2012example}.


\section{Related works}

\subsection{ML meets IRT}

The work \cite{prudencio2015analysis} seek to take the first steps to employ IRT in ML, the aim of this work is to understand the relationship between a dataset considered difficult and the performance of the models. Where they consider that once they get the knowledge that a given classifier performs better in datasets with instances considered difficult, this makes this method preferable over the others. This analysis is compared to the methodology used for psychometric analysis of the proficiency level of students on a test, using the IRT.

In this study, several Random Forests models with different numbers of trees were used to generate the set of responses to estimate the item parameters. For a case study, the Heart-Statlog dataset and the two-parameter logistic model (2PL) were used, focusing on the difficulty parameter. In addition, the work also uses IRT to identify instances considered noise through the intentional insertion of false instances. To compare the performance of classifiers, from the calculation of the hit probability, three different classifiers were used: Naive Bayes, Logistic Regression and Random Forests.

Another work that also employs IRT in ML is \cite{martinez2016making}. In this work, the objective is also to apply IRT as a method to understand how different classification algorithms behave when faced with difficult instances of a dataset. In addition to trying to verify if the so-called difficult instances are actually more difficult than the others or if they are just noise. Furthermore, it also seeks to provide an overview of IRT and how it can be used to resolve the many issues that exist about machine learning.

This work is a continuation of the work mentioned above, its main differences are the use of several classifiers from 15 families of algorithms to generate the set of answers. As a case study, they use the Cassini and Heart-Statlog datasets. In addition to proposing the use of artificial classifiers to serve as a baseline between optimal and bad classification in a linear way. This time, the three-parameter logistic model (3PL) was chosen. In addition to presenting the concept of Classifier Characteristic Curve (CCC) as a way to visualize and analyze the variation in the classifiers' performance on instances with different values of difficulty and discrimination.

Martínez-Plumed et al. (2019) \cite{martinez2019item} is the most complete work, as it aims to describe a pipeline of how to apply IRT in machine learning experiments and explores the advantages of its use, with a focus on supervised learning. In the work, the authors discuss how each item parameter can be used to carry out a deeper analysis about the result of the classifiers. In addition, it is also observed the difference in the use of different logistic models of the IRT, where the 3PL presents the most consistent results.

To perform the IRT analyses, this study used a set of 12 real datasets plus an artificial dataset. In addition, 128 classifiers from 11 different algorithm families were used. The objective is to explore why instances have different item parameter values and how this affects the performance of various learning algorithms. At the end of the work, the authors also suggest five main areas of ML in which IRT can be applied, they are: using IRT to improve classifiers; creation of portfolios of algorithms; classifier selection; improve understanding of the complexity of datasets; and evaluation of classifiers using IRT.

Like the present work, \cite{martinez2018dual} use the IRT to assess benchmarks according to the difficulty and discrimination estimators, but unlike the other works mentioned above, the focus is on reinforcement learning instead of supervised learning. The authors use the benchmarks Arcade Learning Environment (ALE) \cite{bellemare2013arcade} and General Video Game AI (GVGAI) \cite{perez20152014}, which are remarkable benchmarks that allow observing the performance of AI agents in different problems. In the work, dual indicators are proposed to evaluate both the benchmarks and the AI agents, coming from different reinforcement learning algorithms, using the IRT concepts and proposing the generality indicator. Which can be understood as the ability of an agent to solve all tasks up to a certain level of difficulty, which can be limited by available resources.

The authors apply the IRT 2PL logistic model for dichotomous items. For this, they use the human performance threshold in each analyzed game, where: if the AI agent's performance can equal or surpass the human, then it is considered the correct answer, otherwise it is an incorrect answer. In addition to benchmarking, the authors also use IRT's estimated ability and generality to assess agents. In order to use IRT to create the most suitable benchmarks, by selecting games with high discrimination values and accurately measuring whether the AI agent is really generalizing or is specializing in specific tasks.

The use of IRT applied to AI is not limited to evaluating datasets and classification models. Recent work has also used IRT in other contexts in the Machine Learning universe, such as in \cite{de2024explanations}, where the authors use IRT as a new Explainable Artificial Intelligence (XAI) approach to generate global explanations about tree-based models. This work presents a methodology that applies the concepts of IRT psychometrics to explain model predictions and create global rankings of feature relevance. The final results were compared with tools from the XAI literature, showing agreement for the most relevant features.

Another work that also uses IRT in a different context is that of \cite{araujo2023quest}, which focuses on using IRT as a new way of evaluating models within the MLOps paradigm. This work aims to add a new layer in the evaluation of models within the MLOps process with the objective of evaluating whether the proposed model is actually ready to be used in the real world, i.e., whether the model really managed to generalize or whether it only achieves good results during the training.

\subsection{Benchmarking}

In addition to OpenML-CC18 made available by OpenML, other works also highlight the importance of creating and maintaining good benchmarks, such as \cite{nie2019adversarial}. In this work the authors propose a new benchmark for NLI (Natural Language Inference), in which the benchmark is developed using an iterative human-and-model-in-the-loop adversary procedure. In this format, humans first write problems that models cannot classify correctly. The resulting new hard instances serve to reveal model weaknesses and can be added to the training set to create stronger models. Therefore, the new model undergoes the same procedure to collect weaknesses in several rounds, where after each cycle a new stronger model is trained and a new set of tests is created. This cycle can be repeated endlessly to create stronger models and harder benchmarks after each iteration.

Based on this premise, Facebook launched Dynabench \cite{facebook}, a platform for dynamic data collection and benchmarking. The goal is to use the adversary method to iteratively create SOTA (state of the art) models and benchmarks, so you can create a benchmark that doesn't get outdated over time.


Like the studies presented above, this work also seeks to use IRT as a tool for analyzing datasets and classifiers. Among the objectives of this work, we seek to evaluate the well-known benchmark OpenML-CC18 according to the IRT lens, in order to explore its evaluation ability. Alongside this, it is proposed to use the Glicko-2 rating system in conjunction with IRT as a new strategy to perform a more robust assessment of a classifier's strength and to assess the quality and efficiency of subsets of a benchmark. 
As well as the use of the discrimination parameter to filter and choose which games would be more suitable to compose a specific benchmark is similar to the strategy adopted in this work to create more efficient benchmarks. And like Dynabench \cite{nie2019adversarial}, this work aims to create and maintain quality benchmarks, evaluating their ability to test classifiers through IRT parameters.

\section{Materials and methods}

This work proposes the use of Glicko-2 rating system to summarize the data generated by the IRT and define a final score that is capable of measuring the classifiers' ability and also evaluate the quality of benchmarks. Given the fact that rating systems are widely used to measure an individual's ability in an activity, where rating is the numerical value that measures the ability \cite{vevcek2014chess}. With the IRT-Glicko combination, the goal is to use IRT's instance-level evaluation along with Glicko's skill measurement robustness to evaluate datasets and classifiers at local and global levels.

\begin{figure}[htbp]
\centering
\includegraphics[width=1 \textwidth]{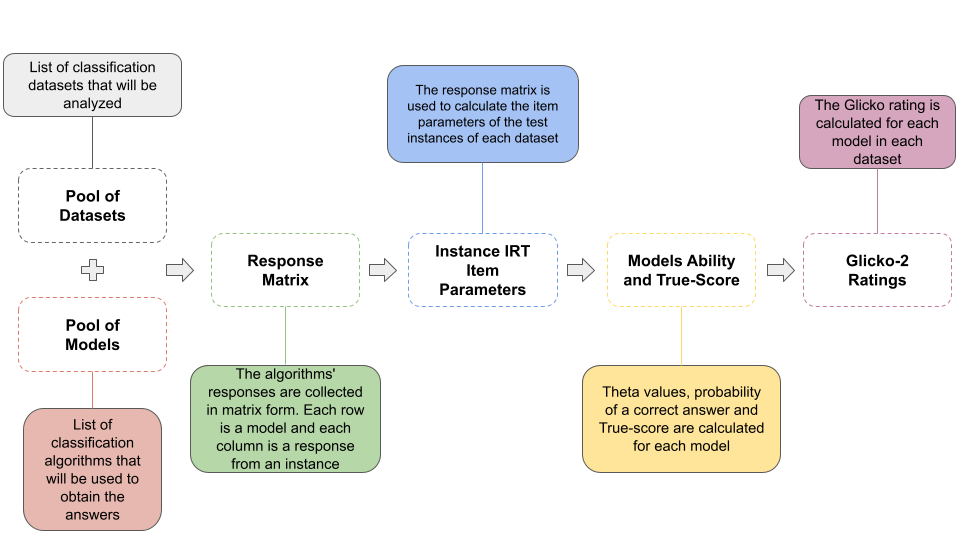}
\caption{Flowchart of the proposed methodology.}
\label{fig:flowchart}
\end{figure}

Figure \ref{fig:flowchart} presents an overview of the proposed methodology. As input we have the set of datasets (benchmark) and the set of models that we wish to analyze. The result of the classification of each of the models for each of the datasets is used to generate the response matrix that is used to estimate the item parameters. Next, the ability of the models is estimated and the True-Score is calculated. Finally, the True-Score values are used to calculate the ratings of each model with the Glicko-2 system. The results generated by the IRT-Glicko combination are then used to evaluate both the datasets and the models. More details about the methodology steps will be presented in the following subsections.

\subsection{Fitting an IRT model}
\label{fitting_irt_model}

Although generally applied for educational purposes, IRT has recently been extended to AI and more specifically to ML \cite{prudencio2015analysis,martinez2016making,martinez2019item}. For this, the following analogy is used: the datasets are the tests, the instances of a dataset are the items and the classifiers are the respondents. In this way, it is as if each test instance were a question from a multiple choice test that the model is answering, where the alternatives are the possible classes of the dataset.

Despite the existence of three IRT logistic models, the 3PL was used for this work, considered by \cite{martinez2019item} as the most complete and consistent logistic model. Furthermore, as stated in previous sections, for the other two logistic models the guessing is always considered equal to 0. However, it is not possible to state that a model cannot correctly classify an instance by luck. Therefore, it is understood the great importance of maintaining the guessing parameter to calculate the probability of correctness of the models.

The item parameters are then used to evaluate the datasets directly, reporting the percentage of difficult instances, with great discriminative power and with a great chance of random hits. In this way, it is possible to have a view of the complexity of the evaluated datasets and how different classifiers behave in the challenge of classifying different datasets.

To calculate the probability of correct answer, you must first estimate the item parameters and the ability of respondents. According to \cite{martinez2016making}, there are three possible situations. In the first, only the item parameters are known. In the second situation, only the ability of the respondents is known. And in the third, and also the most common case, both the items parameters and the respondents ability are unknown. This work is in the third case and for this situation, the following two-step interactive method proposed by Birnbaum \cite{birnbaum1968statistical} is applied:

\begin{itemize}
  \item At first, the parameters of each item are calculated only with the answers of each individual. Initial respondent capability values can be the number of correct answers obtained. For classifiers, this study used the accuracy obtained as the initial ability.
  
  \item Once obtained the items parameters, the ability of individuals can be estimated. For both item parameters and respondent ability, simple estimation techniques can be used, such as maximum likelihood estimation \cite{martinez2016making}.
\end{itemize}

\subsection{Running IRT for ML}

The IRT estimation consists of a total of three main steps designed to be followed in sequence. The first step (Classification and Response Matrix) is responsible to obtain datasets, generating the ML models and placing them to classify the datasets. Then, a response matrix is generated, which contains the classification result of all classifiers for each test instance. The response matrix is the input to the second step (Item Parameter Estimation), which in turn is responsible for calculating the item's parameters for each dataset. The last step (Dataset and Model Evaluation) will use the data generated by the previous ones to rank the datasets using the item parameters and estimate the ability, calculate the response probability and the True-Score of each model. To build the IRT logistic models and analyze the benchmarks, the decodIRT
tool initially presented in \cite{cardoso2020decoding} was adpted.

\subsubsection{Classification and Response Matrix}
\label{decoirt.otml}

\begin{figure}[!htbp]
\centering
\includegraphics[width=1 \textwidth]{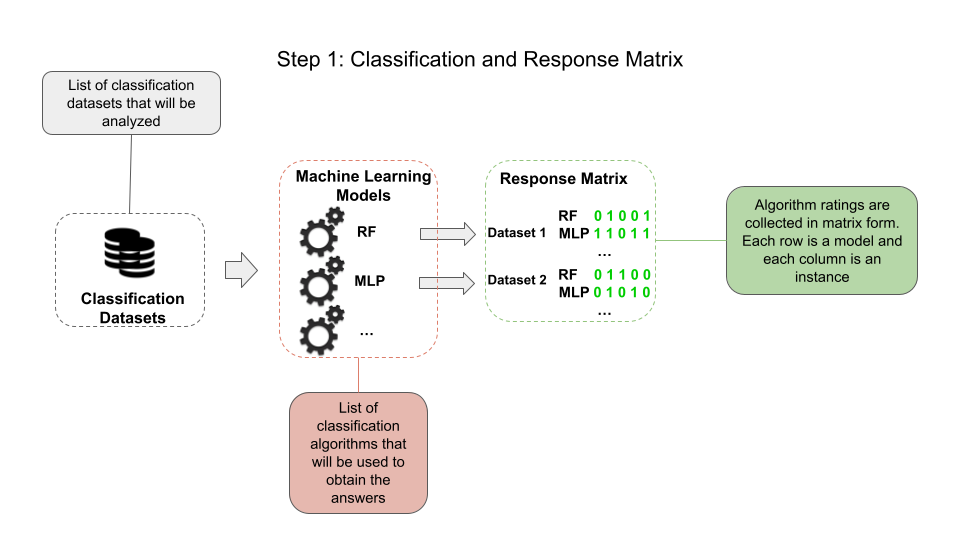}
\caption{Flowchart of the first step.}
\label{fig:model_evaluation}
\end{figure}

The first step has the function of running the models on the datasets to get the answers that are used to estimate the item parameters (see Figure \ref{fig:model_evaluation}). As usual, the datasets are divided into a training set and a test set. So the answers from the classifiers are obtained only for the test set. By definition a stratified split of $70\% \vert 30\%$ is performed, but for very large datasets, the split is handled so that the test set is composed of 500 instances at most. This number of instances is considered large enough for analysis and will be better justified later.

All ML models are generated using Scikit-learn \cite{pedregosa2011scikit} as standard library. Three different sets of classifiers are generated. The first set is composed only of Neural Network (MLP) models. Totaling 120 MLP models, where the depth of the networks gradually increases from 1 to 120. The second set is composed of 12 classifiers from different families that are evaluated in this work, they are: Naive Bayes Gaussian standard, Naive Bayes Bernoulli standard, KNN of 2 neighbors, KNN of 3 neighbors, KNN of 5 neighbors, KNN of 8 neighbors, Standard Decision Trees, Random Forests (RF) with 3 trees, Random Forests with 5 trees, Standard Random Forests, Standard SVM and Standard MLP. The models classified as standard means that the standard Scikit-learn hyperparameters were used. All models are trained using 10-field cross-validation.

The third set of models is composed of 7 artificial classifiers. The concept of artificial classifiers is initially presented in \cite{martinez2016making}, as follows: a optimal classifier (gets all the classifications right), a pessimal one (all misses), a majority (classifies all instances with the majority class), a minority (classify with the minority class) and three random classifiers (sort randomly). This set is used to provide performance threshold indicators for real classifiers. And despite using OpenML as the base repository, decodIRT also allows the user to use local datasets and define training and testing sets specifically.

\subsubsection{Item Parameter Estimation}

\begin{figure}[!htbp]
\centering
\includegraphics[width=1 \textwidth]{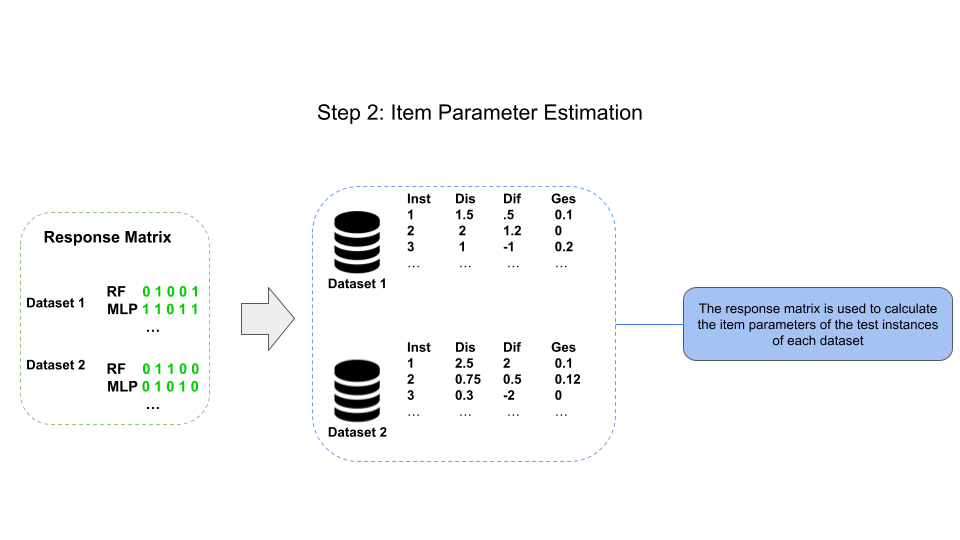}
\caption{Flowchart of the second step.}
\label{fig:item_parameter_estimation}
\end{figure}

The second step has the function of using the responses generated by the classifiers and estimating the item parameters for the test instances (see Figure \ref{fig:item_parameter_estimation}). As stated previously, the logistic model for dichotomous items is used, which means that regardless of the number of classes existing in each dataset, it is only considered if the classifier was right or wrong in the classification of each instance.

To calculate the item parameters, the Ltm package \cite{rizopoulos2006ltm} for the R language is used, which implements a framework containing several mechanisms for the calculation and analysis of the IRT. The Rpy2 package \cite{gautier2008rpy2} was used to perform Python communication with the R packages. As mentioned previously, the maximum limit of 500 instances for estimating item parameters was defined. According to \cite{martinez2019item}, packages that estimate the IRT item parameters may get stuck in a local minimum or not converge if the number of items is too large. This is not strange, as the IRT is used for psychometric tests, it is very unusual for these tests to have such a large number of questions. Thus, it is recommended that less than 1000 instances be used to estimate the parameters.

\subsubsection{Dataset and Model Evaluation}

\begin{figure}[!htbp]
\centering
\includegraphics[width=1 \textwidth]{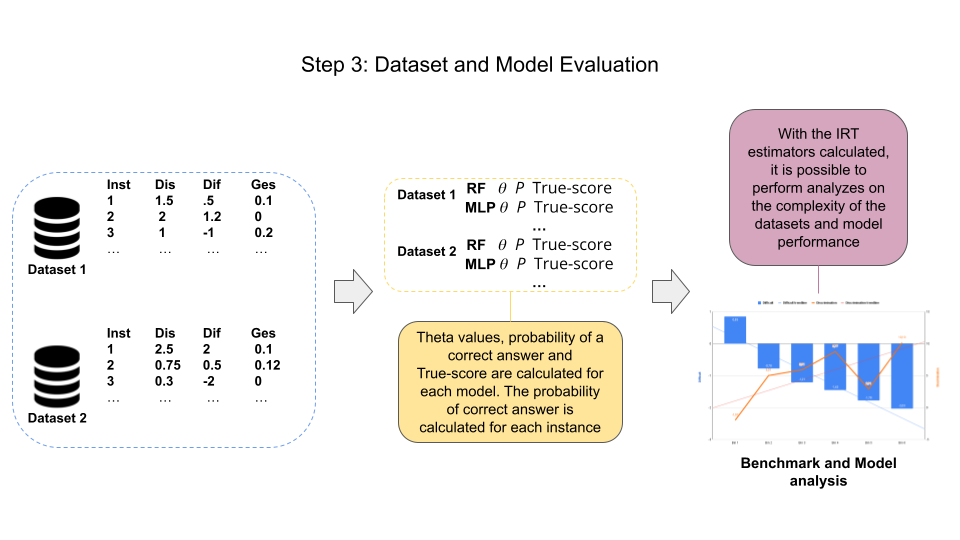}
\caption{Flowchart of the third step.}
\label{fig:dataset_model_evaluation}
\end{figure}

The third step involves carrying out an analysis and organizing the data generated by the previous steps, in order to make the data easier to read (see Figure \ref{fig:dataset_model_evaluation}. Among the various functions of this step is the creation of dataset rankings by item parameter. 

Before calculating the probability of correct answer of the classifiers for the instances, one must first estimate the ability of the classifiers, as explained in Birnbaum's method. Both to estimate the ability $\theta$ and to calculate the probability of correct answer, use the Catsim package \cite{meneghetti2017application} from Python. For this, the instances are sorted according to their difficulty and divided into 10 groups, then they are used in ascending order to estimate the ability of the classifiers. After this step, the probability of a correct answer can be calculated to obtain the True-Scores that will be used to generate the Glicko ratings.

\subsection{Ranking of classifiers by the Glicko-2 system}
\label{glicko_met}

Due to the fact that rating systems are commonly used in competitions, to apply the Glicko-2 \cite{glickman2012example} system to evaluate the classifiers, it was necessary to simulate a competition between them. The simulated competition is an round-robin tournament, where each classifier will face each other and at the end of the competition will create a ranking with the rating of the models.

\begin{figure}[htbp]
\centering
\includegraphics[width=1 \textwidth]{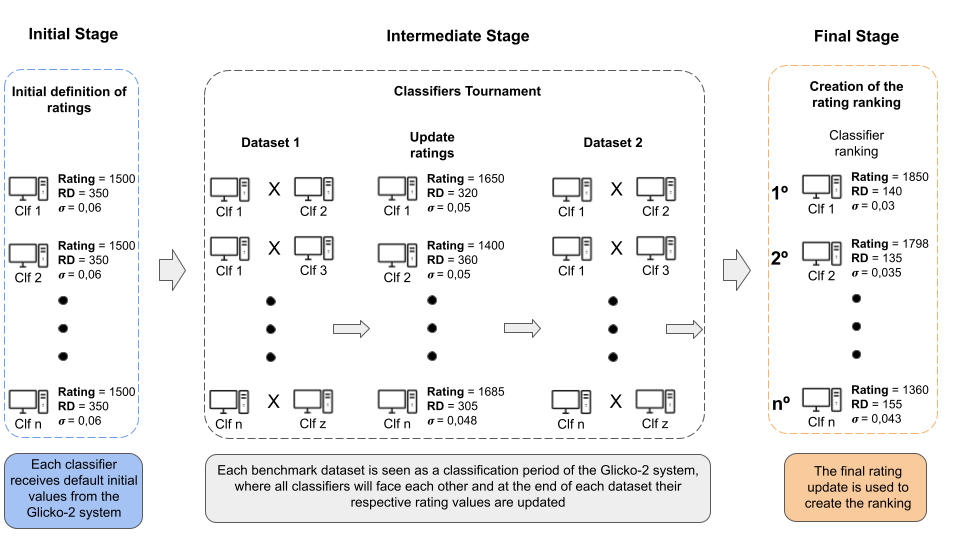}
\caption{Example of ratings generation by the Glicko-2 system.}
\label{fig:glicko}
\end{figure}

The competition works like this: each dataset is seen as a classification period in the Glicko-2 system, so that all classifiers face off in each dataset. To define the winner of each contest, the True-Score values obtained by the models facing each other are used. This happens as follows: if the True-Score value is greater than the opponent's, it is counted as a victory; if the value is lower than that of the opponent, it is counted as a defeat; and if the values are equal, then it sets up a tie between the models. In addition, the Glicko system asks the result of a match to assign a score to the opponents. For this, the scoring system applied in official Chess competitions was used, where victory counts as 1 point, defeat as 0 point and draw counts as 0.5 point.

Thus, after each dataset the rating, RD and volatility values of the classifiers are updated and used as the initial value for the next dataset. Once all datasets are finalized, the final rating values are used to create the final ranking that will be used to evaluate the models.

\subsection{Interpreting IRT for ML}
\label{interpreting_IRT_ML}


Given the multidisciplinary nature of the proposed methodology, before presenting the results obtained, this subsection aims to explain the interpretation of the IRT estimators when applied to ML.

As presented in subsection \ref{fitting_irt_model} in this study, the IRT 3PL model is used, when considering the concept of each of the item parameters, it is understood that their values can be interpreted as follows:

The difficulty parameter $b_{i}$ is the simplest to interpret. The higher its value, the more difficult the instance will be to classify correctly. This means that there is some characteristic within the instance that makes it more difficult than the others. The increase in difficulty may be related both to the structure of the instance itself, such as a missing value or outlier of a feature, or to its own conception when defining the problem.

The discrimination parameter $a_{i}$ defines how well an instance can be used to separate low and high skill models. So, for an instance with a high discrimination value, models that classify it correctly probably have high skill, while models that classify it incorrectly probably have low skill. It can be considered that more discriminatory instances are more suitable for evaluating the performance of classifiers. The discriminatory quality of an instance is possibly related to its own design, so a dataset with well-assembled instances tends to have higher discrimination values.

The guessing parameter $c_{i}$ represents the minimum probability that any model has of classifying the instance correctly. Its value is a probability that varies between 0 and 1, and the higher its value means that less skill is required to correctly classify the instance. So, even low-skill models can have a high chance of getting the classification right. Each instance can present a different guessing value, it is understood that this parameter can be useful to indicate shortcut learning conditions, where the instance has information that by chance induces the correct classification \cite{cardoso2022explanation}.

In this work, item parameters are explored at the dataset level, as the focus is on the study of model evaluation using benchmarks. Works such as \cite{martinez2019item,cardoso2022explanation} study the relationship of item parameters at the instance level.

\section{Results and discussion}

OpenML-CC18 was chosen to be the case study of this work, which has as one of the main objectives to evaluate benchmarks through the IRT lens, in order to give greater reliability in the use of this benchmark. This section will present the datasets that were selected from OpenML-CC18 to be evaluated using the decodIRT tool.

Despite having 72 datasets, only 60 were used in this work. This was for two main reasons:

\begin{enumerate}
    \item The size of the datasets, where 11 have more than 30,000 instances, were soon considered too large and would take a long time to run all decodIRT models;
    \item Could not generate item parameters for dataset ``Pc4''. R's Ltm package could not converge even using just under 500 test instances.
\end{enumerate}

Despite this, the final amount of datasets used still corresponds to 83.34\% of the original benchmark. All datasets evaluated are from tabular data and the characterization of the benchmark will be further explored in the next sections.

The evaluation of the OpenML-CC18 benchmark through the IRT lens was done around the discrimination and difficulty parameters. It is understood that these parameters are directly linked to the data, in comparison with the guessing parameter that is more linked to the performance of the respondents. The objective then is to evaluate the discriminatory power along with the difficulty of the datasets and later use them to evaluate the models performance, in addition to exploring the impact of using different datasets to evaluate classifiers.

\subsection{Decoding OpenML-CC18 Benchmark}
\label{results_1}

\begin{figure}[!ht]
\centering
\includegraphics[width=1 \textwidth]{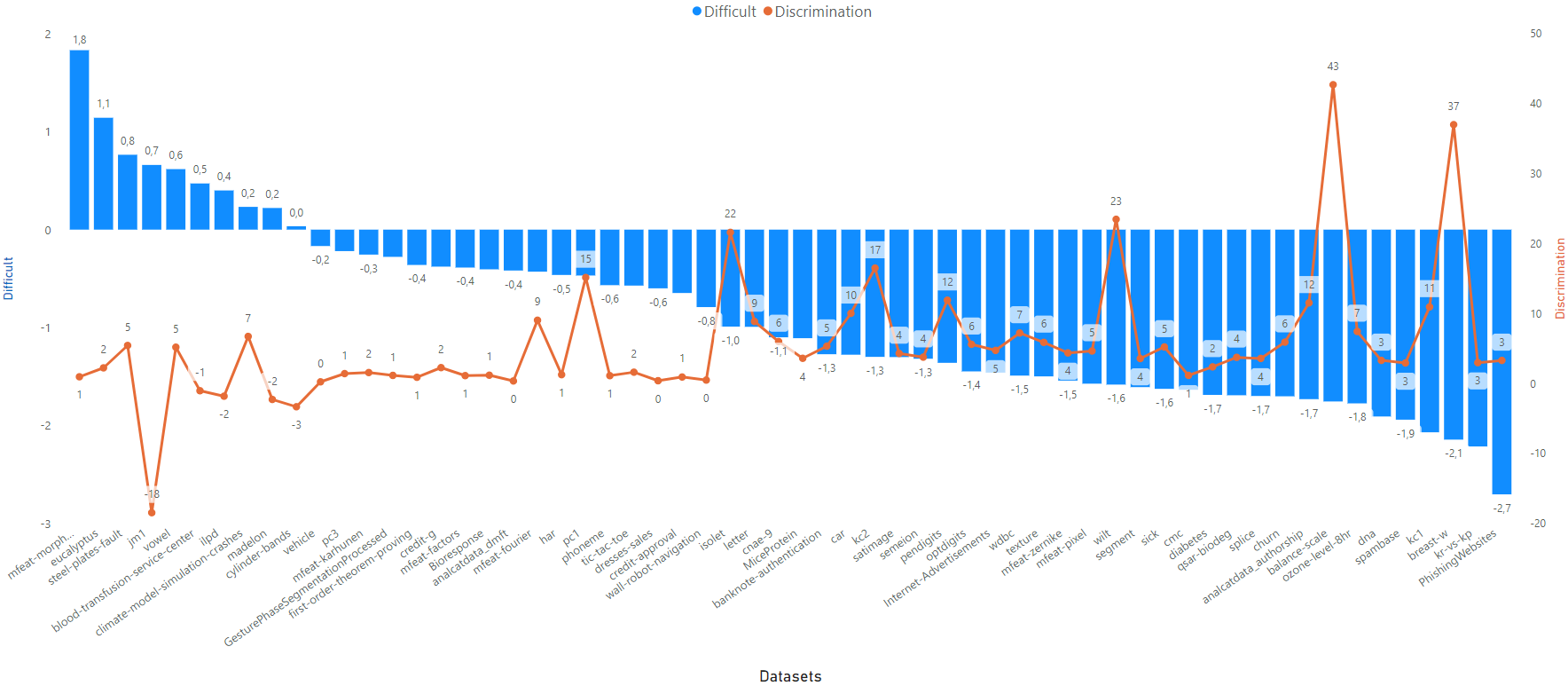}
\caption{Difficulty and Discrimination values obtained for the benchmark ordered by Difficulty. The blue columns are the mean difficulty per dataset, while the orange line is the mean discrimination.}
\label{fig:dis_dif}
\end{figure}

Once the proposed experiments were carried out, the first step was to understand whether there is any relationship between the item parameters obtained from each benchmark dataset. Figure \ref{fig:dis_dif} presents the comparison between the mean values obtained for the difficulty and discrimination parameters for each dataset. Firstly, it is noted that the benchmark is composed of datasets with different levels of difficulty and discrimination. This means that within the benchmark there are different sub-sets of easier or more challenging datasets, for example. At first glance, it can be seen that only 16.66\% of the benchmark has difficulty values above 0 and only 3.33\% have difficulty values above 1. This means that only 1/6 of the datasets are truly challenging.

\begin{figure}[!bp]
\centering
\includegraphics[width=.8 \textwidth]{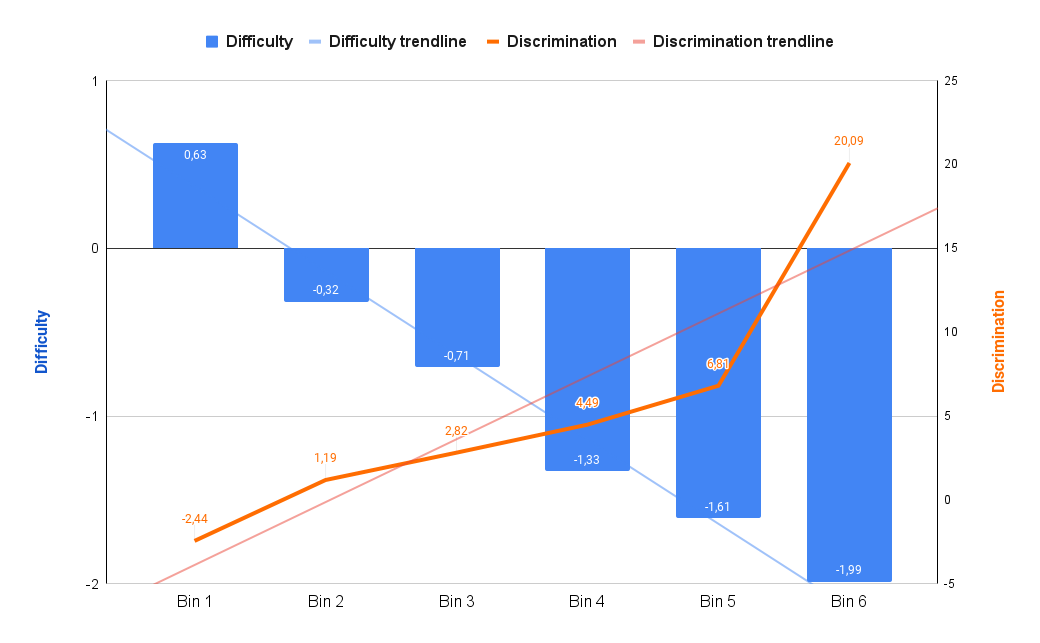}
\caption{Difficulty and Discrimination values obtained for each bin ordered by Difficulty. The blue columns are the mean bin difficulty, while the orange line is the mean discrimination.}
\label{fig:dis_dif_trendline}
\end{figure}

Regarding discrimintation, more than 90\% of the benchmark datasets present positive discrimination values and would be good for separating highly skilled and unskilled classifiers.
It is noted that there is a tendency for the discrimination and difficulty values to have an inverted relationship (see Figure \ref{fig:dis_dif}).
To better evaluate this condition, the benchmark was divided into 6 equal bins of 10 datasets, the first being composed of the most difficult and the last of the easiest and then the average of the discrimination values of each set was calculated. The result is that the most difficult bin with 0.63 difficulty has -2.44 discrimination, while the easiest bin with -1.99 difficulty has 20.09 discrimination. Figure \ref{fig:dis_dif_trendline} makes this relationship more evident, it is possible to notice that the trend line for discrimination is rising while the trend line for difficulty is falling. The complete list of separate datasets in each Bin can be consulted in Table \ref{tab_datasets}.

\begin{table}[!ht]
\fontsize{7.3pt}{7.3pt}\selectfont
\begin{center}
\caption{OpenML-C18 Datasets.}\label{tab_datasets}%
\begin{tabular}{@{}llccc@{}}
\toprule
Bins & Dataset & Difficulty & Discrimination & Guessing\\
\midrule
\multirow{10}{*}{Bin 1} & mfeat-morphological               & 1.83        & 0.967         & 0.02       \\
                        & eucalyptus                        & 1.14        & 2.24          & 0.03       \\
                        & steel-plates-fault                & 0.76        & 5.44          & 0.02       \\
                        & jm1                               & 0.66        & -18.5         & 0.02       \\
                        & vowel                             & 0.61        & 5.20          & 0.02       \\
                        & blood-transfusion-service-center  & 0.47        & -1.02         & 0.00       \\
                        & ilpd                              & 0.4         & -1.81         & 0.00       \\
                        & climate-model-simulation-crashes  & 0.23        & 6.73          & 0.00       \\
                        & madelon                           & 0.22        & -2.27         & 0.00       \\
                        & cylinder-bands                    & 0.034       & -3.32         & 0.00       \\ 
                        \midrule
\multirow{10}{*}{Bin 2} & vehicle                           & -0.16       & 0.24          & 0.05       \\
                        & pc3                               & -0.22       & 1.42          & 0.02       \\
                        & mfeat-karhunen                    & -0.25       & 1.57          & 0.03       \\
                        & GesturePhaseSegmentationProcessed & -0.27       & 1.16          & 0.05       \\
                        & first-order-theorem-proving       & -0.36       & 0.89          & 0.02       \\
                        & credit-g                          & -0.37       & 2.28          & 0.00       \\
                        & mfeat-factors                     & -0.38       & 1.13          & 0.07       \\
                        & Bioresponse                       & -0.40       & 1.19          & 0.08       \\
                        & analcatdata\_dmft                  & -0.41       & 0.37          & 0.01       \\
                        & mfeat-fourier                     & -0.43       & 9.05          & 0.01       \\
                        \midrule
\multirow{10}{*}{Bin 3} & har                               & -0.46       & 1.27          & 0.06       \\
                        & pc1                               & -0.46       & 15.2          & 0.06       \\
                        & phoneme                           & -0.56       & 1.13          & 0.07       \\
                        & tic-tac-toe                       & -0.57       & 1.62          & 0.08       \\
                        & dresses-sales                     & -0.59       & 0.40          & 0.04       \\
                        & credit-approval                   & -0.64       & 0.93          & 0.01       \\
                        & wall-robot-navigation             & -0.79       & 0.49          & 0.10       \\
                        & isolet                            & -0.99       & 21.6          & 0.00       \\
                        & letter                            & -0.99       & 8.87          & 0.01       \\
                        & cnae-9                            & -1.1        & 6.01          & 0.03       \\
                        \midrule
\multirow{10}{*}{Bin 4} & MiceProtein                       & -1.11       & 3.61          & 0.05       \\
                        & banknote-authentication           & -1.27       & 5.34          & 0.12       \\
                        & car                               & -1.28       & 10.0          & 0.02       \\
                        & kc2                               & -1.3        & 16.5          & 0.00       \\
                        & satimage                          & -1.3        & 4.3           & 0.08       \\
                        & semeion                           & -1.32       & 3.77          & 0.04       \\
                        & pendigits                         & -1.36       & 11.9          & 0.01       \\
                        & optdigits                         & -1.45       & 5.61          & 0.02       \\
                        & Internet-Advertisements           & -1.46       & 4.74          & 0.05       \\
                        & wdbc                              & -1.49       & 7.23          & 0.03       \\
                        \midrule
\multirow{10}{*}{Bin 5} & texture                           & -1.5        & 5.88          & 0.01       \\
                        & mfeat-zernike                     & -1.54       & 4.38          & 0.01       \\
                        & mfeat-pixel                       & -1.57       & 4.65          & 0.03       \\
                        & wilt                              & -1.58       & 23.5          & 0.05       \\
                        & segment                           & -1.61       & 3.57          & 0.10       \\
                        & sick                              & -1.63       & 5.23          & 0.06       \\
                        & cmc                               & -1.64       & 1.17          & 0.04       \\
                        & diabetes                          & -1.69       & 2.38          & 0.03       \\
                        & qsar-biodeg                       & -1.69       & 3.74          & 0.08       \\
                        & splice                            & -1.7        & 3.59          & 0.11       \\
                        \midrule
\multirow{10}{*}{Bin 6} & churn                             & -1.7        & 5.92          & 0.01       \\
                        & analcatdata\_authorship            & -1.73       & 11.5          & 0.02       \\
                        & balance-scale                     & -1.75       & 42.7          & 0.07       \\
                        & ozone-level-8hr                   & -1.77       & 7.44          & 0.01       \\
                        & dna                               & -1.91       & 3.31          & 0.09       \\
                        & spambase                          & -1.94       & 2.94          & 0.08       \\
                        & kc1                               & -2.07       & 11.0          & 0.03       \\
                        & breast-w                          & -2.14       & 37.0          & 0.02       \\
                        & kr-vs-kp                          & -2.21       & 2.99          & 0.11       \\
                        & PhishingWebsites                  & -2.70       & 3.30          & 0.08     \\
\bottomrule
\end{tabular}
\end{center}
\end{table}

This relationship is consistent with what is expected by the IRT, where it is normal that the easiest instances are good to differentiate the good from the bad classifiers, as it is thought that the more skilled classifiers will hit the easiest instances while the less skillful ones can make mistakes. Through this, it is possible to affirm that the more difficult datasets are not adequate to separate the good and bad classifiers, despite being more challenging. Meanwhile, the easiest datasets are not suitable for testing the classification power of algorithms, but it allows to differentiate the wrost algorithms from the minimally adequate.

Despite the IRT estimators, defining whether an instance is difficult or not is still a challenge, as the concept of difficulty itself is subjective.
The best case would be to find items that are both very discriminative and difficult, but this is not a simple task given the observed inversion relationship. In Figure \ref{fig:dis_dif_trendline} it is noted that only Bin 2 could partially fit these ideal conditions, as it presents the second highest average of difficulty and positive discrimination.

But why doesn't Bin 1 work? Since it has the highest difficulty? Despite the high difficulty value, Bin 1 presents the problem of negative discrimination. As they are not expected by the IRT, negative discrimination values usually mean that there is something wrong with the item itself. For psychometric tests, this could mean a poorly formulated and ambiguous question, for example. When placing this concept in the ML field, negative discrimination may indicate some inconsistency in the instance, such as noise or outlier. Therefore, it can be inferred that datasets with many instances with negative discrimination may not be suitable for the formulation of a good benchmark.


\begin{figure}[!tbp]
  \centering
  \subfloat[With negative discrimination.]{\includegraphics[width=0.5\textwidth]{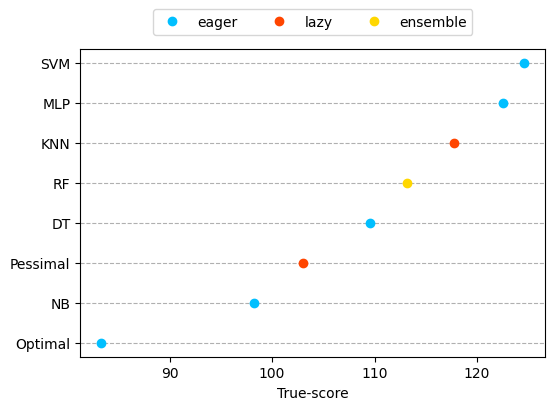}\label{fig:f1}}
  \hfill
  \subfloat[No negative discrimination.]{\includegraphics[width=0.5\textwidth]{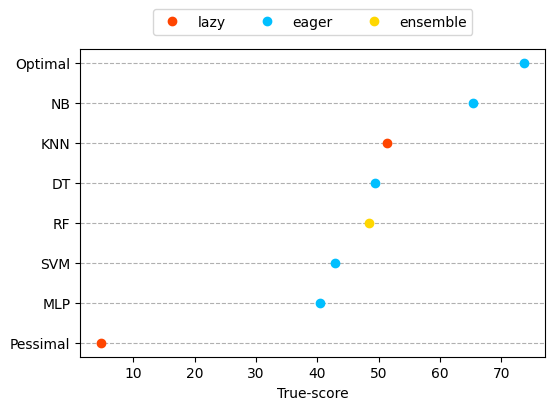}\label{fig:f2}}
  \caption{The True-Score values obtained for the ``ilpd'' dataset. The values in Figure (a) were calculated using all instances, including those with negative discrimination. While the values in Figure (b) were calculated without the instances with negative discrimination.}
  \label{fig:negative_dis}
\end{figure}

For example, Figure \ref{fig:negative_dis} presents the True-Score values obtained for the data set \textit{``ilpd''} belonging to Bin 1, where it can be seen that the pessimal artificial classifier obtained a higher score than the optimal classifier that obtained the worst score when considering all instances (see Figure \ref{fig:f1}), a situation that should not happen. Note that when considering only instances with positive discrimination, the Optimal and Pssimal classifiers respectively present the best and worst True-Score, as expected (see Figure \ref{fig:f2}). Such a condition may reveal a data set that is not concise and may not be suitable for a benchmark, including model explanation/interpretation. A possible future work would be to carefully analyze whether the characteristics of the dataset are linked to these situations and how this can affect the performance of the models. OpenML already has an extensive set of metadata about its datasets that can be used for this purpose..

The \textit{``ilpd''} dataset (see Figure \ref{fig:negative_dis}), for example, is a binary dataset with more than 70\% of its instances belonging to the majority class, such imbalance could be one of the causes of negative discrimination.
In this dataset there are 416 records of liver patients and 167 records of non-liver patients, in addition, among these records 441 are male and 142 are female. The number of records of female patients for the \textit{``hepatic''} class is only 22.11\% while for the \textit{``non-hepatic''} class it is only 29.94\%. This means that the dataset has another imbalance within its instances that can harm the models learning.



It is also understood that the creation of a benchmark depends on the purpose for which you wish to evaluate the models, in this sense the IRT estimators can help select the datasets most suitable for the purpose. For example, if the objective is solely to test the classification power of an algorithm, choosing a set of difficult datasets would be more appropriate. On the other hand, if the objective is to evaluate the interpretability of a model, a more discriminative benchmark may be more appropriate.

During the experiments, it was found that some datasets present high variations between the average values of each item parameter. For example, within the same dataset there are instances with very high and very low values of the same item parameter. For a benchmark, depending on the set of instances divided for training and testing, the performance of a model can vary greatly, which can impair the direct comparison of results. Despite this, such information can still be useful for carrying out adapted training, where the model would be trained with the easiest instances and then tested with the most difficult ones and vice versa, for example.

\begin{figure}[htbp]
\centering
\includegraphics[width=1 \textwidth]{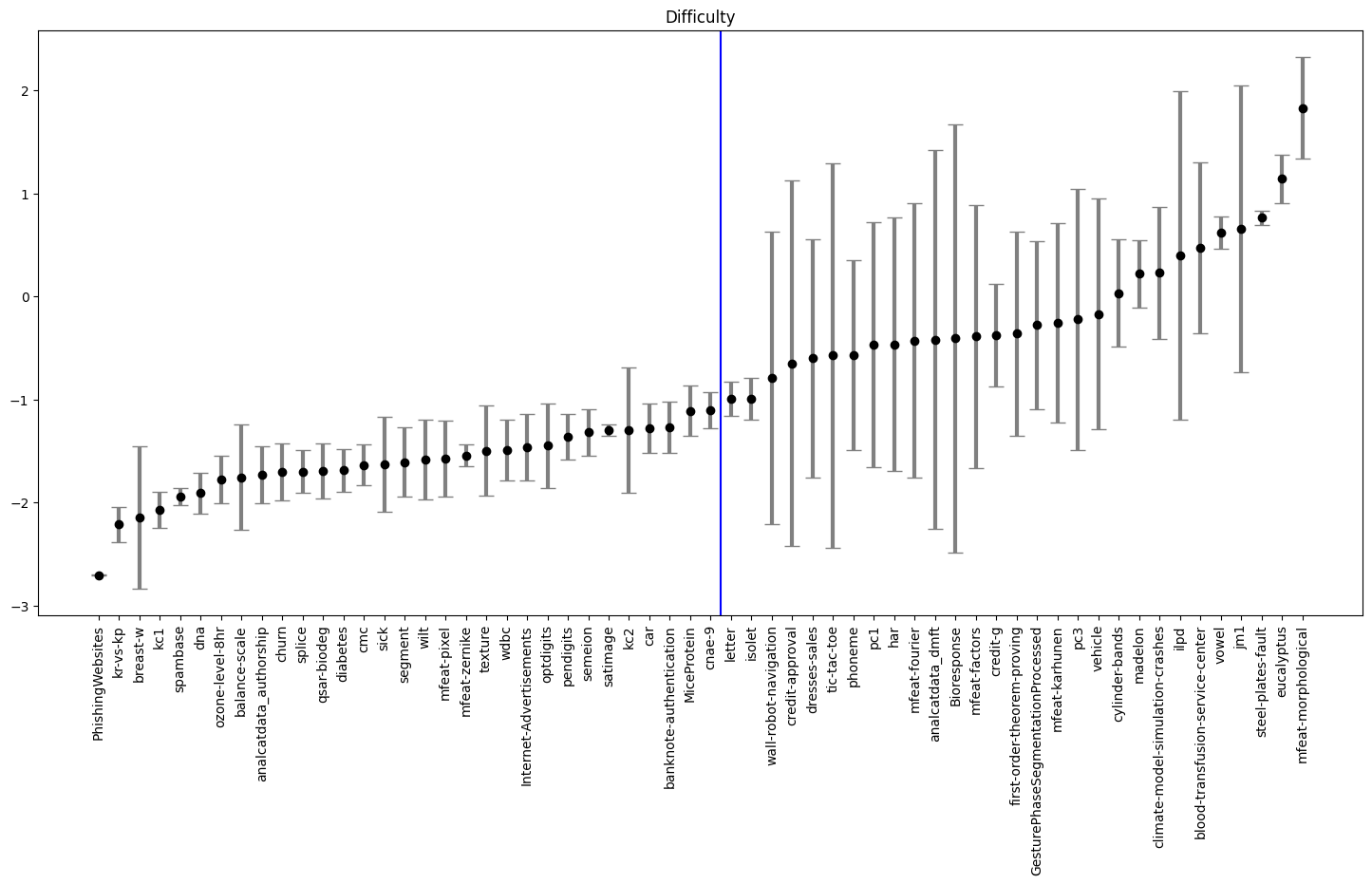}
\caption{Standard deviation of difficulty of instances of each Dataset.}
\label{fig:desvio_dis}
\end{figure}

In fact, it is not surprising that in a classification task there are simpler and more complex cases to classify, but this condition can be more harmful than useful since classic ML metrics do not consider the complexity of the item. 
There is no consensus on whether a dataset with high or low variation in the difficulty level of its instances is preferable. 
With this in mind, the OpenML-CC18 datasets were separated by the Standard Deviation obtained for each parameter, in order to allow choosing those that present the smallest variations. When analyzing Figure \ref{fig:desvio_dis} it is clear that there is a tendency for the most difficult datasets to also have the greatest variations and although this is not a rule, the same also happens for the other item parameters. 
In this way, it is possible to interpret that a given dataset can be considered challenging not only due to the nature of the problem, but also due to the inconsistency in the construction of the dataset itself, as an example there is the \textit{``ilpd''} observed dataset previously.


Although models and data are two sides of the same coin, it is common knowledge that a model does not exist if the data does not previously exist. When proposing a classification task, the first step is to clearly define the problem and what data to use. A dataset that has such high variations in the item parameter values of its instances may be an indication that the dataset was not formed properly, therefore it would not be preferable to be part of a benchmark.

\begin{figure}[htbp]
\centering
\includegraphics[width=1 \textwidth]{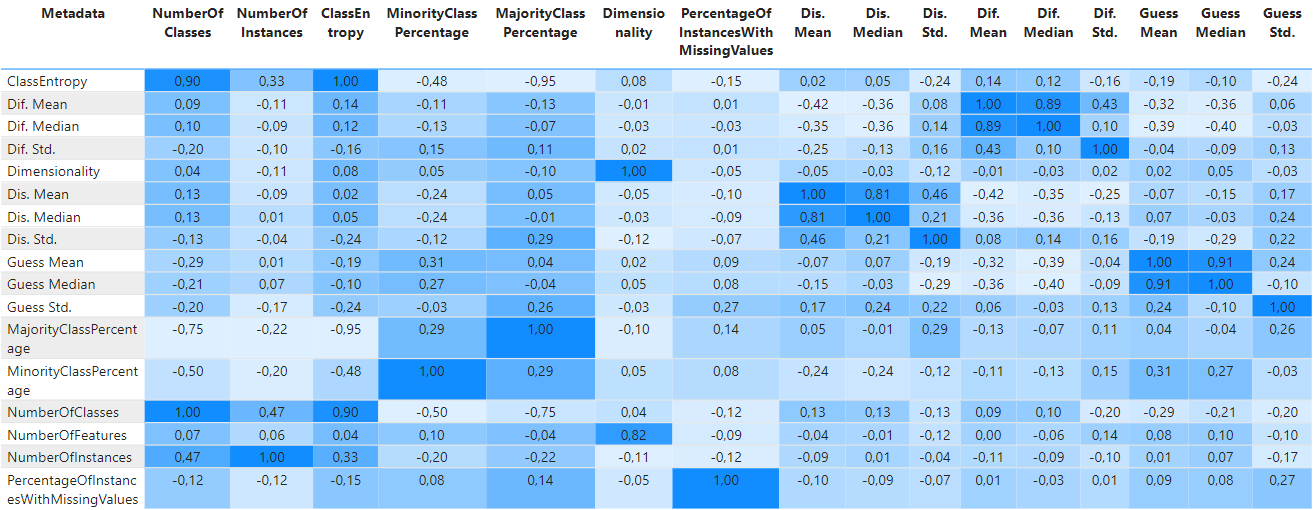}
\caption{Correlation between dataset metadata and item parameters.}
\label{fig:matriz_corr}
\end{figure}

This leads to the hypothesis mentioned previously: \textit{Can the general characteristics of datasets indicate when a dataset will have a high or low value for a given item parameter?} \textit{For example, could imbalance in a dataset be indicative of an item parameter?}
Therefore, this relationship was explored using as a basis seven metadata from each dataset obtained from OpenML itself, they are: \textit{NumberOfClasses}, \textit{NumberOfInstances}, \textit{NumberOfFeatures}, \textit{ClassEntropy}, \textit{Dimensionality}, \textit{PercentageOfInstancesWithMissingValues}, \textit{MajorityClassPercentage} and \textit{MinorityClassPercentage}. This metadata was chosen because it is information that describes any dataset in general. When analyzing the correlation obtained (see Figure \ref{fig:matriz_corr}), it is observed that the most relevant correlations were:

\begin{itemize}
    \item \textit{NumberOfClasses} with average Guessing (-0.29) and Difficulty deviation (-0.20);
    \item \textit{PercentageOfInstancesWithMissingValues} with guessing deviation (0.27);
    \item \textit{MajorityClassPercentage} with Guessing deviation (0.26) and Discrimination deviation (0.29);
    \item  \textit{MinorityClassPercentage} with average Guessing (0.31) and average Discrimination (-0.24); 
\end{itemize}

The correlation values obtained make sense when interpreting the relationship between the concept of the item parameter and the concept of the dataset metadata itself. For example, for a multiclass dataset it is expected that the classes will have different levels of classification difficulty, which directly impacts the Difficulty standard deviation value. The relationship between the number of classes and the Guessing average is also coherent, because if a dataset has many classes it is not surprising that the models were not able to generalize adequately and started to guess the classification. As well as the correlation between Guessing and the percentage of instances with missing values, as the lack of information about an instance can force models to guess its class.

The percentages of the minority and majority classes being more strongly correlated with the Guessing and Discrimination parameters is also consistent, as it is expected that the average guessing value will be higher for the minority class. Discrimination, in turn, presents a negative correlation with the percentage of the minority class. This makes sense, as the minority class is traditionally the most difficult class for the model to get right, so it tends to have lower and even negative discrimination values, as seen in the example of the ``ilpd'' dataset in Figure \ref{fig:negative_dis}.

Another point to highlight is that ClassEntropy presents a high correlation with all Standard Deviation values of item parameters. This information makes sense, since Class Entropy represents the degree of class disorder, it is expected that datasets with high Entropy values will present large disagreements between item parameters.

A possible future work would be to use the instance-centric properties of IRT to explore the correlations of the instances' own features with the item parameters.

\subsection{Classifiers performance by IRT and Glicko-2}

This section aims to explore the performance of the classifiers according to the rating values obtained by Glicko-2. But before analyzing the experiments, it is important to answer some questions. It has already been exemplified in Section \ref{irt_section} that IRT can present different information than classic evaluation metrics, so that respondents with the same number of correct answers can present different scores depending on the items they answered correctly. So why apply the Glicko-2 rating system if the IRT itself is already used to evaluate?

Like the IRT, the rating system also aims to measure an individual's ability, but there are differences in the abilities that are measured. It is common for chess players to use the word ``strength'' when talking about someone's rating, and perhaps this is the best word to describe the type of ability that rating systems measure. Strength has several meanings and comes from the Latin \textit{``fortia''}, among which it can be understood as the ability to fulfill a certain task, very much like a great demonstration of power. Rating systems like Glicko-2 aim to measure this great demonstration of strength or skill within a given activity. IRT, in turn, is commonly used in proficiency tests that aim to measure the depth of an individual's knowledge about a certain area and their competence in it. For example, an individual with a lot of knowledge in Chess may not be very strong at playing while an individual who is very strong at playing may not know how to explain the reason for his strength.

In ML it is very common to just look at the strength of the models, if the model impresses with high power (accuracy) it is often enough for that model to be preferable. By applying the Glicko-2 rating system to the classifiers at the IRT, the aim is to carry out a fairer assessment of the real ability of the classifiers and try to measure strength without disregarding proficiency.


As already explained in Section \ref{glicko_section}, the Glicko system measures strength based on a competition between players, in this case between classifiers. To this end, a round-robin style competition was held between all classifiers (details in Section \ref{glicko_met}). In rating systems, the order of confrontation also influences the growth or decline of the rating obtained. For example, if you start the evaluation with the easiest datasets and gradually advance to the more difficult ones, it may generate a different final ranking if the opposite is done.

To define the order and set of datasets used in the evaluation of Glicko, the results of the IRT estimators presented in the previous subsection were used. To evaluate the validity of the chosen set of datasets, the performance of the artificial classifiers will be used as a baseline. For example, very controversial orders such as Pessimal's high performance and Optimal's low performance will be considered, as they may represent an unsuitable benchmark set.

There are countless combinations of datasets that can be made from the IRT estimators to form a benchmark from the 60 OpenML-CC18 datasets, but it is not feasible to compare all possible combinations. Although it is not a simple task, one of the objectives of this work is to verify whether it is possible to select a subset of datasets that is fairer to evaluate the models considering the information obtained by IRT. Therefore, this work focuses on testing different subsets of data considering mainly the balance and the average level of difficulty and discrimination of the datasets.


6 different sets of datasets were defined to evaluate the classifiers:

\begin{itemize}
\item All 60 datasets ordered from least difficult to most difficult.
     \item All 60 datasets ordered from least discriminative to most discriminative.
     \item The 30 datasets with the smallest variation in the difficulty parameter, ordered from smallest deviation to largest deviation.
     \item The 30 datasets with the greatest variation in the difficulty parameter, ordered from smallest deviation to largest deviation.
     \item The 30 datasets with the smallest variation in the discrimination parameter, ordered from smallest deviation to largest deviation.
     \item The 30 datasets with the greatest variation in the discrimination parameter, ordered from smallest deviation to largest deviation.
\end{itemize}

For simplicity, only 12 classifiers were considered to generate the Glicko rating rankings, the 6 artificial ones described in Section \ref{decoirt.otml}, along with the real classifiers: SVM, MLP, RF, KNN, NB and DT.  


\begin{table}[h]
\begin{center}
\caption{Classifier rating ranking with increasing discrimination.}\label{tab1}%
\begin{tabular}{@{}clccc@{}}
\toprule
Rank & Classifier & Rating & RD & Volatility\\
\midrule
1 & optimal                                 & 1799.001 & 39.37 & 0.061     \\ 
2 & SVM                                   & 1734.361 & 36.899 & 0.06     \\ 
3 & MLP                         & 1700.412 & 36.607 & 0.06     \\ 
4 & RandomForest                & 1697.411 & 37.094 & 0.06    \\ 
5 & KNeighbors               & 1690.538 & 36.496 & 0.06     \\ 
6 & GaussianNB              & 1525.258 & 36.402 & 0.062     \\ 
7 & DecisionTree                            & 1511.353 & 36.018 & 0.061     \\ 
8 & majority                           & 1423.922 & 36.639 & 0.062     \\ 
9 & rand2                                 & 1257.671 & 37.688 & 0.06     \\ 
10 & rand1                                 & 1225.113 & 37.679 & 0.06     \\ 
11 & minority                           & 1204.294 & 41.085 & 0.06    \\ 
12 & pessimal                               & 1064.439 & 42.822 & 0.062     \\
\bottomrule
\end{tabular}
\end{center}
\end{table}

As can be seen in Table \ref{tab1}, it was the Optimal classifier that obtained the highest rating with almost 60 pts more than the second placed (SVM). The order of artificial classifiers ended as expected, Optimal in first and the other artificial ones in the last positions, with Majority being the best among them and Pessimal being the worst. Furthermore, there is a difference of almost 90 points between the last real classifier (DT) and the majority. For the real classifiers, it was SVM that obtained the highest rating with more than 30 pts ahead of MLP and RF, respectively. The two worst performances of the real classifiers were NB and DT, with a very close rating, being the only real classifiers that remained at around 1500 points.

When compared to the Difficulty ranking (see Table \ref{tab2}), the most striking differences are the huge decline of the Optimal classifier, with 170 pts less difference in relation to the Discrimination ranking, falling to 3rd place, and the Pessimal classifier which showed an increase of around 290 pts in the classification, rising to the penultimate place.


\begin{table}[h]
\begin{center}
\caption{Classifier rating ranking with increasing difficulty.}\label{tab2}%
\begin{tabular}{@{}clccc@{}}
\toprule
Rank &Classifier & Rating & RD & Volatility\\
\midrule
1 & RandomForest                             & 1709.253 & 36.322 & 0.06    \\ 
2 & SVM                 & 1708.006 & 37.202 & 0.06    \\ 
3 & optimal                                 & 1628.72 & 35.712 & 0.061     \\ 
4 & KNeighbors                         & 1612.051 & 35.628 & 0.06     \\ 
5 & MLP               & 1611.458 & 36.552 & 0.06    \\ 
6 & DecisionTree              & 1591.749 & 35.243 & 0.061     \\ 
7 & GaussianNB                            & 1550.593 & 35.602 & 0.063     \\ 
8 & majority                           & 1473.321 & 35.622 & 0.062     \\ 
9 & rand2                                 & 1363.977 & 36.348 & 0.06     \\ 
10 & rand1                                 & 1349.186 & 36.643 & 0.06     \\ 
11 & pessimal                               & 1291.399 & 38.425 & 0.06     \\
12 & minority                           & 1148.405 & 41.288 & 0.06    \\ 
\bottomrule
\end{tabular}
\end{center}
\end{table}

\begin{figure}[!htbp]
\centering
\includegraphics[width=1 \textwidth]{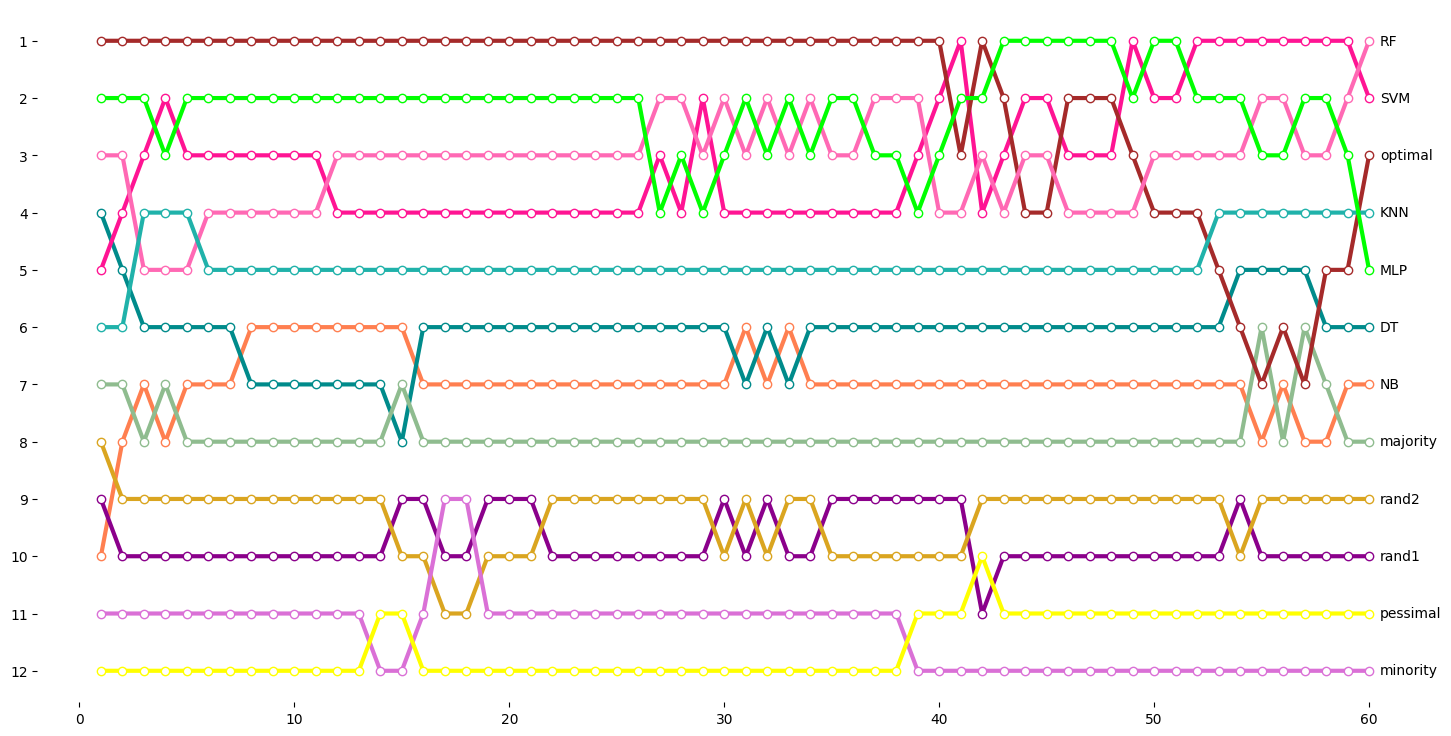}
\caption{Ranking behavior with increasing difficulty.}
\label{fig:bump_dif_menor_maior}
\end{figure}

The explanation for this result is the way the rating system works and the complexity of the datasets from the last rounds of confrontation. As can be seen in Figure \ref{fig:dis_dif} there are a total of 10 datasets that do not present negative Difficulty values and were considered the most difficult in the benchmark, so in ordering by difficulty these are the last 10 datasets used to perform the tournament and estimate ratings. To allow for adequate visualization, a Bump Chart was generated (see Figure \ref{fig:bump_dif_menor_maior}) that shows the ranking of each of the classifiers throughout the tournament. It can be seen that the Optimal classifier led throughout the competition, but began to lose many rating points in the last remaining rounds (datasets). This is due to the high incidence of instances with negative discrimination values, which can also be seen in Figure \ref{fig:dis_dif}, as classifiers with high ability such as Optimal were evaluated with items that favor respondents with low ability such as Pessimal, an improvement in the performance of Pessimal is also noted at the same time that the decline of Optimal begins.

It is understood that the rankings reflect the impact that different item parameter values have on the same classifiers. To create a good benchmark, choosing only the most discriminative datasets or only the most difficult ones may not be the best option. According to \cite{smith2014reducing}, not all instances of a dataset are equally beneficial for learning. This means that some instances can help more than others in the model induction process, and it is even possible to have instances that hinder learning. By expanding this idea to a higher level, it is possible to imagine that in a benchmark, not all datasets are equally beneficial for evaluating models.

The remaining 4 sets of datasets were also constructed taking this into account. They are subsets of the original benchmark, as only half of the total amount of datasets is used to carry out the tournament. The separation was initially done to evaluate the impact of datasets with high variability of item parameter values on their instances and whether this could actually be detrimental to a fair evaluation of models with a benchmark. The results obtained show that sets with the lowest item parameter variation generated the most consistent Glicko rankings.

\begin{table}[h]
\begin{center}
\caption{Classifier rating ranking with less variation in difficulty.}\label{tab3}%
\begin{tabular}{@{}clccc@{}}
\toprule
Rank &Classifier & Rating & RD & Volatility\\
\midrule
1 & optimal                                   & 1975.373 & 46.916 & 0.06    \\ 
2 & RandomForest                & 1808.96 & 39.763 & 0.06    \\ 
3 & MLP                                 & 1762.407 & 39.1 & 0.06     \\ 
4 & SVM                         & 1743.663 & 38.126 & 0.06     \\ 
5 & KNeighbors               & 1667.219 & 37.554 & 0.06    \\ 
6 & DecisionTree              & 1560.567 & 37.498 & 0.06     \\ 
7 & GaussianNB                            & 1557.156 & 37.448 & 0.06     \\ 
8 & majority                           & 1377.26 & 38.232 & 0.06     \\ 
9 & rand2                                 & 1269.044 & 39.331 & 0.06     \\ 
10 & rand1                                 & 1228.674 & 39.679 & 0.06     \\ 
11 & minority                               & 1066.613 & 44.793 & 0.06     \\
12 & pessimal                           & 947.97 & 47.962 & 0.06    \\ 
\bottomrule
\end{tabular}
\end{center}
\end{table}

When observing Table \ref{tab3} one can immediately notice the increase in the range of re rating values. The Optimal classifier remains first, but now with almost 170 pts of difference above the second place, which this time is RF, followed by MLP and SVM. For RF to achieve 1st place, it would be necessary for a maximum negative variation of the Optimal rating to occur within its RD along with a maximum positive variation of the RF's RD. It is also noted that this final ranking agrees in almost every way with the ranking of the complete benchmark presented in Table \ref{tab1}. Furthermore, the increase in the distance between the real NB classifier and the artificial Majority classifier by almost 200 pts stands out, in this case even the maximum fluctuation of the ratings within their respective RD would still not be possible for the Majority to reach the real NB classifier.

\begin{table}[h]
\begin{center}
\caption{Classifier rating ranking with less variation in discrimination.}\label{tab34}%
\begin{tabular}{@{}clccc@{}}
\toprule
Rank &Classifier & Rating & RD & Volatility\\
\midrule
1 & optimal                                   & 1904.183 & 48.278 & 0.06    \\ 
2 & RandomForest                & 1772.354 & 40.504 & 0.06    \\ 
3 & MLP                                 & 1719.655 & 39.501 & 0.06     \\ 
4 & SVM                         & 1676.283 & 38.378 & 0.06     \\ 
5 & KNeighbors               & 1605.714 & 37.915 & 0.06    \\ 
6 & DecisionTree              & 1589.154 & 37.873 & 0.06     \\ 
7 & GaussianNB                            & 1560.558 & 38.08 & 0.061    \\ 
8 & rand2                           & 1359.058 & 39.753 & 0.06     \\ 
9 & rand1                                 & 1353.447 & 39.884 & 0.06     \\ 
10 & majority                                 & 1311.237 & 39.739 & 0.06     \\ 
11 & minority                               & 1067.861 & 44.993 & 0.06     \\
12 & pessimal                           & 1053.999 & 47.406 & 0.06    \\ 
\bottomrule
\end{tabular}
\end{center}
\end{table}

And unlike the rankings generated using the complete benchmark, the rankings obtained using the datasets with the lowest variation in Difficulty and Discrimination agree in the final position of all real classifiers as can be seen in Table \ref{tab34}, the variation was only for the final position of the artificial classifiers where the Majority was surpassed by the random classifiers. Although still high, the difference between the Optimal classifier decreased to around 130 pts between the RF which remains in second. Furthermore, it was observed that the sets of datasets with the lowest variation in difficulty and discrimination agree on 76.6\% of the benchmark as they present only 23 identical datasets.

\begin{figure}[!ht]
\centering
\includegraphics[width=1 \textwidth]{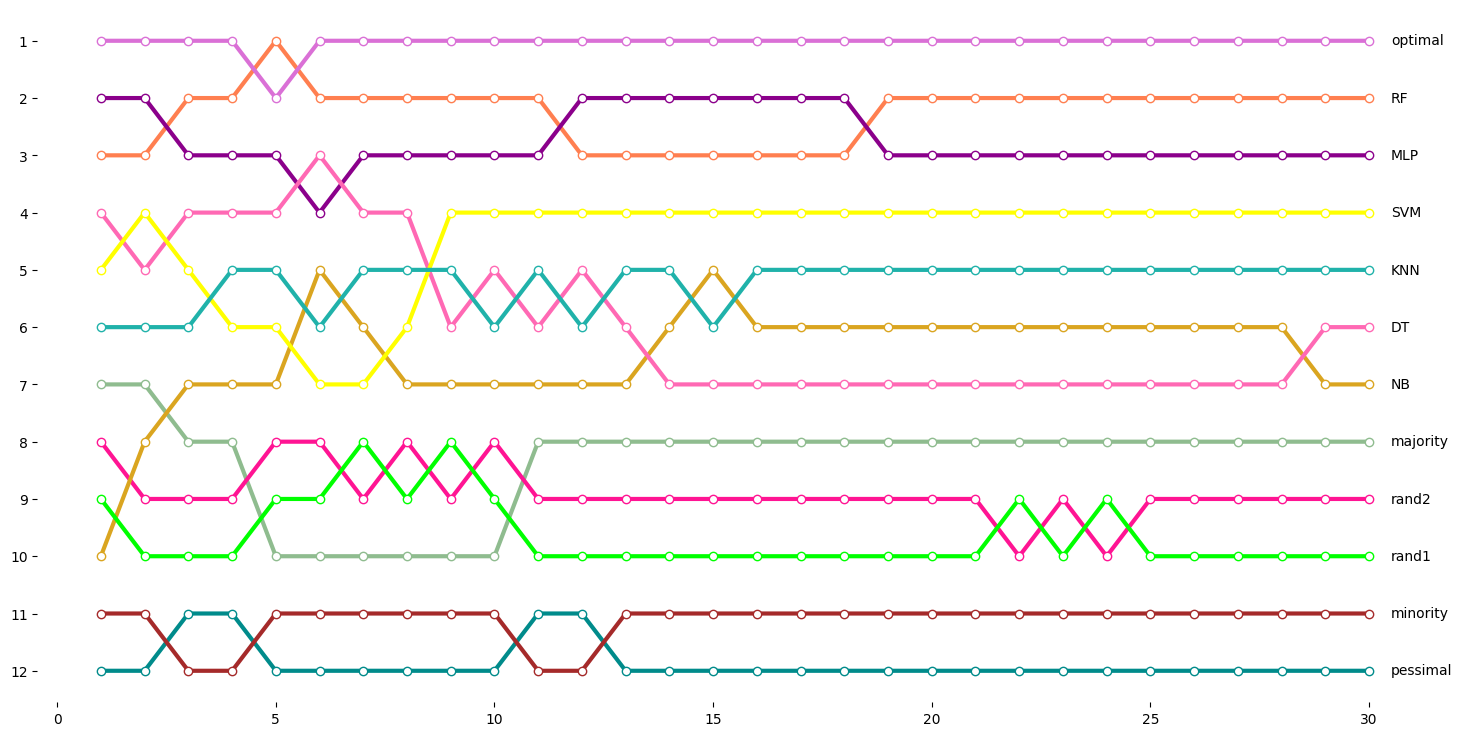}
\caption{Ranking behavior with the lowest difficulty variation.}
\label{fig:bump_dis_menor_desvio}
\end{figure}

The Figure \ref{fig:bump_dis_menor_desvio} allows the behavior of the classifiers to be observed in relation to the proposed benchmark subset. It is noted that for the last 10 datasets (tournament rounds) there was only once a change in the final position of the real classifiers (DT and NB). The clear separation of the classifiers highlights the level of skill that each model possesses, and in this case it was the RF that demonstrated the greatest ability and classification strength.

\begin{table}[h]
\begin{center}
\caption{Classifier rating ranking with greater difficulty variation.}\label{tab4}%
\begin{tabular}{@{}clccc@{}}
\toprule
Rank &Classifier & Rating & RD & Volatility\\
\midrule
1 & SVM                                   & 1716.493 & 37.694 & 0.06    \\ 
2 & MLP                & 1715.031 & 36.47 & 0.06    \\ 
3 & KNeighbors                                 & 1644.446 & 35.867 & 0.06     \\ 
4 & RandomForest                         & 1636.772 & 35.582 & 0.06     \\ 
5 & DecisionTree               & 1553.809 & 34.936 & 0.061    \\ 
6 & optimal              & 1538.281 & 35.378 & 0.061     \\ 
7 & majority                            & 1473.67 & 35.306 & 0.062     \\ 
8 & GaussianNB                           & 1439.852 & 35.194 & 0.062     \\ 
9 & rand2                                 & 1336.638 & 36.653 & 0.06     \\ 
10 & pessimal                                 & 1333.571 & 36.932 & 0.06     \\ 
11 & rand1                               & 1309.886 & 36.556 & 0.06     \\
12 & minority                           & 1152.307 & 39.894 & 0.06    \\ 
\bottomrule
\end{tabular}
\end{center}
\end{table}

Contrary to the consistent results from the subsets with less variation in item parameters, the rankings generated by the datasets with greater variation presented very questionable results (see Table \ref{tab4}). As can be seen, the Optimal classifier is only in 6th position in the ranking and Pessimal is in 10th ahead of the minority. RF, previously presented as the strongest classifier, had a lower final performance than KNN.



\subsection{Relationship between dataset metadata and models}

As seen in Subsection \ref{results_1} and Figure \ref{fig:matriz_corr}, it is understood that different characteristics of the datasets indicate a correlation with the item parameters estimated by the IRT. This section aims to explore how this relationship can influence the performance of different ML algorithms.

For this, the 60 evaluated datasets were ordered according to the metadata taken from OpenML, for example, datasets with a lower percentage of minority class to the datasets with a higher percentage. As in Figure \ref{fig:dis_dif_trendline}, the datasets were combined into 6 bins of 10 datasets and for each bin the average IRT probability of correct response was calculated. 

\begin{figure}[!htbp]
\centering
\includegraphics[width=1 \textwidth]{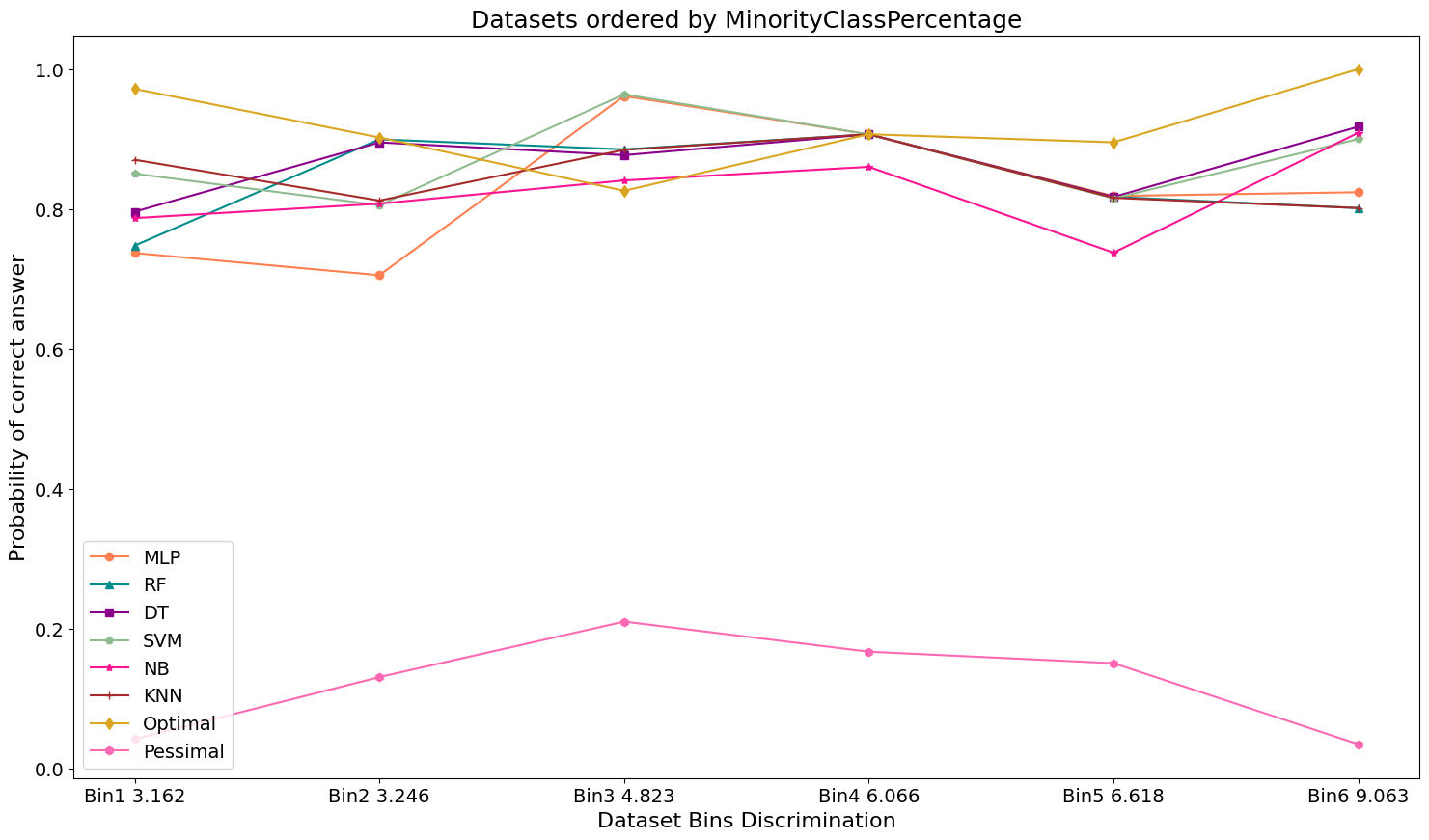}
\caption{Probability of correct answer with increasing Minority Class Percentage.}
\label{fig:class_crescente_prob}
\end{figure}

Figure \ref{fig:class_crescente_prob} illustrates the relationship between the variation of datasets with an increasing minority class percentage and the probability of success of different ML models. There is a growing increase in the average level of discrimination as the percentage of minority classes also increases. As previously observed in subsection \ref{results_1}, for Bin 1 where the minority class has a low frequency in the datasets the Discrimination tends to be lower, while for Bin 6 which has the minority class closer to the majority class as it is composed of Binary datasets present greater discrimination due to the greater occurrence of minority instances.



\begin{figure}[!htbp]
\centering
\includegraphics[width=1 \textwidth]{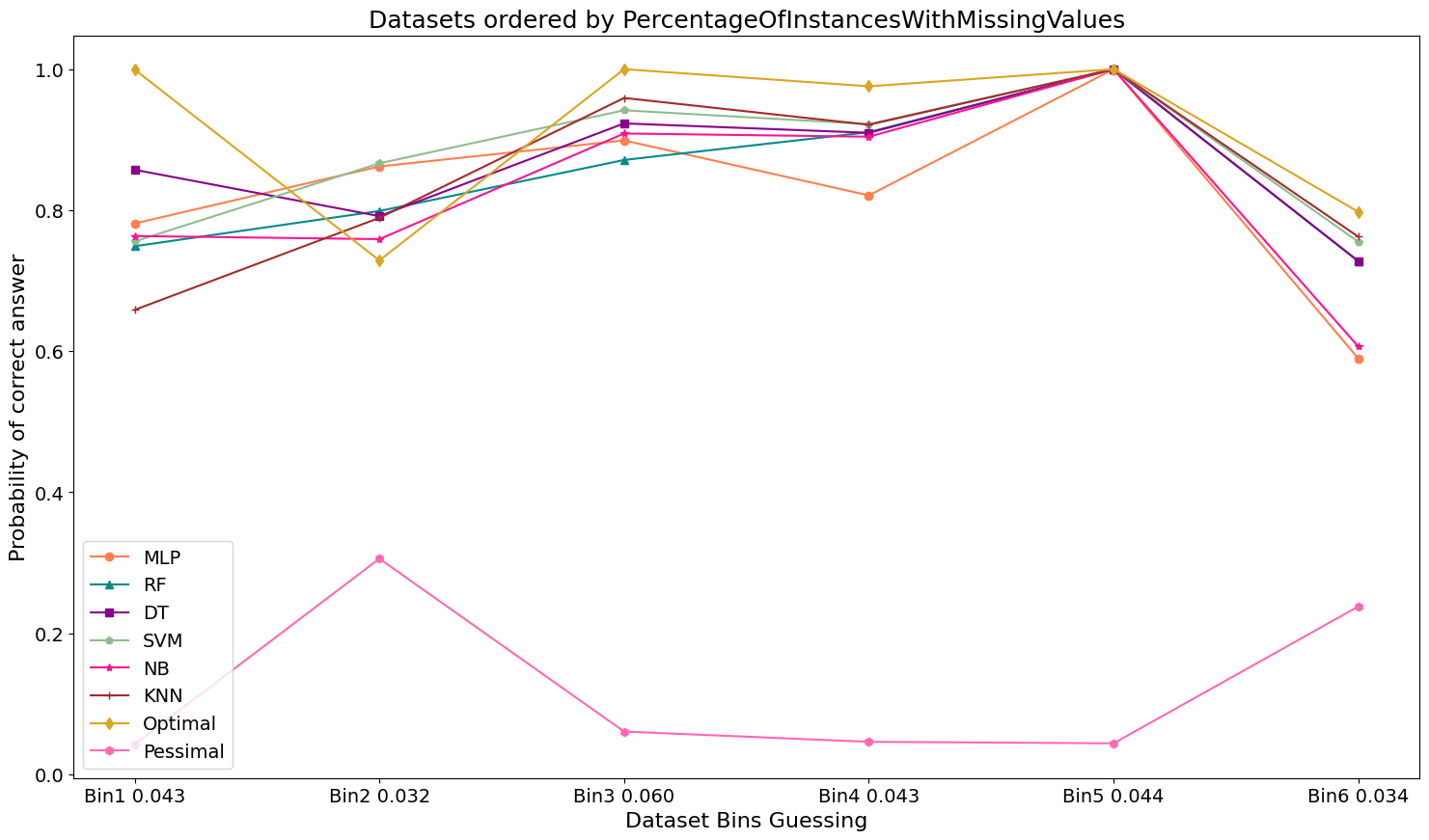}
\caption{Probability of correct answer with increasing Percentage of Instances with Missing Values.}
\label{fig:missing_crescente_prob}
\end{figure}

Figure \ref{fig:missing_crescente_prob} illustrates the relationship between the guessing parameter and the percentage of instances with missing values in the datasets. As already imagined, the number of missing values negatively impacts the performance of the classifiers. So, in the last Bin (set of datasets with the most missing values), the MLP and NB were the most affected models.

The relationship with guessing can be observed directly in the performance obtained by the models in Bins 3 and 5, which are the sets of datasets with the highest guessing values and are also the Bins where the classifiers presented the highest correct answer probability values, this indicates that the missing values in the datasets of these Bins do not negatively impact the final performance of the models, but induce guessing as previously observed at the end of subsection \ref{results_1}.

On the other hand, Bins 2 and 6 that have the lowest guessing values are also where the models presented the lowest performance, and are also the only bins that have a dataset with negative discrimination values. Specifically, Bin 6, which has the lowest average guess value, is also the Bin with the most datasets with missing values, this was already expected, as it is natural that missing values in the datasets make it difficult for the classifier to casually get the instance right. In this case, Bin 6 also has the highest average difficulty value, with -0.581.

\begin{figure}[!ht]
\centering
\includegraphics[width=1 \textwidth]{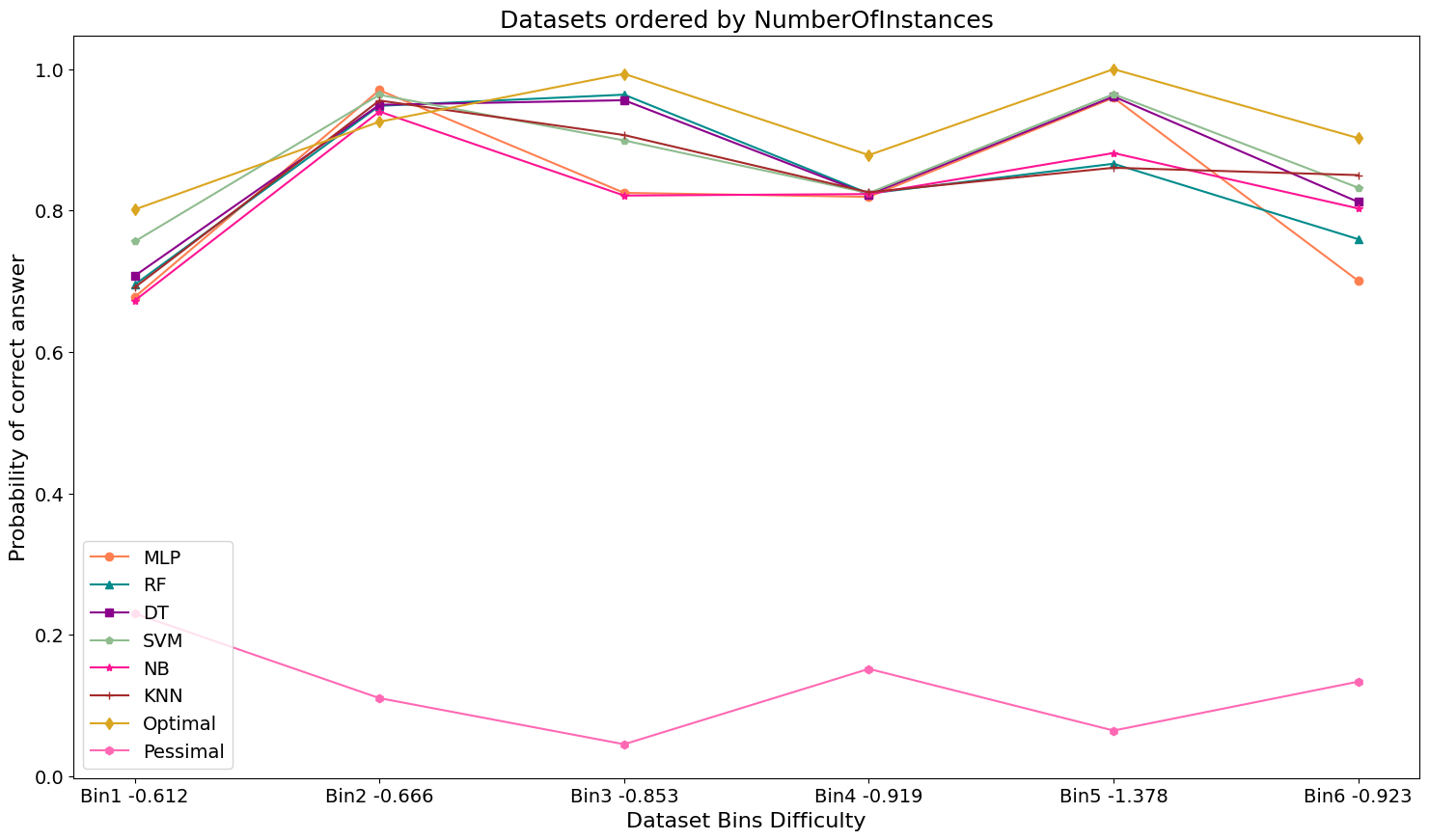}
\caption{Probability of correct answer with increasing Number of Instances.}
\label{fig:instance_crescente_prob}
\end{figure}

The third case observed is the simplest to analyze, as can be seen in Figure \ref{fig:instance_crescente_prob} as the number of instances in the datasets increases, the average difficulty value tends to decrease. This is already expected within ML, algorithms learn inductively through examples. If more examples are provided, it is a natural tendency that the generated models will be better. Despite this, it is interesting to note that for the most difficult Bin 1, it was the SVM classifier that achieved the highest correct response probability, as seen in Table \ref{tab2}. 

It is interesting to note that for Bin 6, which has the largest number of instances per dataset, it was the RF and MLP models that presented the lowest correct answer probability, while KNN presented the highest correct answer probability. A future study would be to analyze the quality of the existing instances in the Bin 6 datasets, and explore whether the large quantity actually reflects the quality of the models' learning.




\section{Final considerations}

This work explored the IRT for benchmark evaluation, that commonly are used to explore how far ML algorithms can go when dealing with distinct datasets before selecting the most stable and strong classifier. Although OpenML-CC18 is designed to be a gold standard, it should be used with caution. Of the 60 datasets evaluated, only 16.66\% have instances that are considered difficult, while 90\% of the of the benchmark presents high discrimination values. This condition can be a great source for analyzing comparisons, but it is not useful for testing the ability of classifiers. The benchmark assessment methodology based on IRT is provided and can be replicated by the decodIRT tool in an automated way.

In addition, IRT and Glicko was used to explore whether there is a more efficient benchmark subset than the original one and whether it could be obtained more information about models ability and data complexity from the IRT estimators. After exploring different subsets, the subset consisting of 50\% of the total datasets selected on the percentages of discrimination and difficulty was chosen. The use of the Glicko rating system with IRT also allows for more meaningful comparisons between classifiers and provides greater confidence in which model performs best on which set of datasets.

The subset proved to be more suitable for evaluating and separating the strength of the models. Furthermore, the creation of benchmark subsets allowed us to further explore the models' abilities. The final result showed that Random Forest is the classifier with the greatest ability, making its choice preferable in relation to the other models evaluated when facing the complexities of different data sets. Therefore, it has been demonstrated that IRT can also be used to filter and create more efficient datasets from benchmarks.

The work also explored the relationship between dataset properties and IRT item parameters. It was observed that certain dataset metadata correlate with Difficulty values, for example. Therefore, the very constitution of a dataset must be considered when choosing a classification algorithm. Furthermore, it was also observed that not all datasets may be suitable for composing a benchmark, with datasets with negative discrimination not being recommended for model evaluation. Such a condition may be associated with the presence of malformed instances within a data set.

Although this study focuses on the relationship between models and data at the dataset level, the analysis performed can be re-done at the local level, considering the performance of models against specific instances of a single dataset. An instance-level study on different types of datasets can be useful to evaluate the impact of malformed instances on model learning and performance. Finally, it is reiterated that data and classifieds are two sides of the same coin and must be considered together when analyzing a model.

\section*{Acknowledgements}

This section aims to recognize and thank the efforts of the co-authors who assisted the MSc. Lucas Cardoso (first author), responsible for planning and conducting this research. To PhD Ronnie Alves, who guided this research from beginning to end and helped with ideas and reviews of the entire work. To Vitor Santos, who has been participating in this journey uniting IRT and ML since the beginning and who helped review the methodology. To José Ribeiro who joined during the journey and contributed with reviews and ideas for experiments for the work. To Regiane Kawasaki, who was also helping from the beginning and was the one who presented the research opportunity and helped with reviewing the text. And to Ricardo Prudêncio, who started the studies with IRT and ML and presented it to the group, also contributed with ideas and reviews of the results.

\section*{Funding}

This section aims to register and also thank the institutions that funded this work.

\begin{itemize}
    \item Federal University of Pará - UFPA: research grant.
    \item Vale Technological Institute - ITV: research grant.
    \item Federal University of Pernanbuco - UFPE: research grant.
    \item Federal Institute of Education, Science and Technology of Pará - IFPA: research grant.
\end{itemize}

\section*{Conflicts of interest/Competing interests}

Both authors and institutions that supported this work declare that there is no conflict of interest.

\section*{Ethics approval}

Does not apply to this work.

\section*{Supplementary Material}

Source code, results, additional information and all data generated by this work can be found in this repository: \url{https://osf.io/wvptb/files/}.

\bibliography{bibliography}

\begin{thebibliography}{33}
\providecommand{\natexlab}[1]{#1}
\providecommand{\url}[1]{\texttt{#1}}
\providecommand{\urlprefix}{URL }
\expandafter\ifx\csname urlstyle\endcsname\relax
 \providecommand{\doi}[1]{doi:\discretionary{}{}{}#1}\else
 \providecommand{\doi}[1]{doi:\discretionary{}{}{}\begingroup \urlstyle{rm}\url{#1}\endgroup}\fi
\providecommand{\bibinfo}[2]{#2}

\bibitem{araujo2023quest}
V.~C.Araujo~Santos, L.Cardoso, R.Alves, The quest for the reliability of machine learning models in binary classification on tabular data, Scientific Reports 13~(1) (2023) 18464.

\bibitem{baker2001basics}
F.~B.Baker, The basics of item response theory, ERIC, 2001.

\bibitem{bellemare2013arcade}
M.~G.Bellemare, Y.Naddaf, J.Veness, M.Bowling, The arcade learning environment: An evaluation platform for general agents, Journal of Artificial Intelligence Research 47 (2013) 253--279.

\bibitem{birnbaum1968statistical}
A.Birnbaum, F.Lord, M.Novick, Statistical theories of mental test scores, Some latent trait models and their use in inferring an examinee’s ability. Addison-Wesley, Reading, MA (1968).

\bibitem{bischl2017openml}
B.Bischl, G.Casalicchio, M.Feurer, F.Hutter, M.Lang, R.~G.Mantovani, J.~N.vanRijn, J.Vanschoren, Openml benchmarking suites and the openml100, stat 1050 (2017) 11.

\bibitem{burnell2023rethink}
R.Burnell, W.Schellaert, J.Burden, T.~D.Ullman, F.Martinez-Plumed, J.~B.Tenenbaum, D.Rutar, L.~G.Cheke, J.Sohl-Dickstein, M.Mitchell, et~al., Rethink reporting of evaluation results in ai, Science 380~(6641) (2023) 136--138.

\bibitem{cardoso2022explanation}
L.~F.Cardoso, J.deS.~Ribeiro, V.~C.~A.Santos, R.~L.Silva, M.~P.Mota, R.~B.Prud{\^e}ncio, R.~C.Alves, Explanation-by-example based on item response theory, in: Brazilian Conference on Intelligent Systems, Springer, 2022, pp. 283--297.

\bibitem{cardoso2020decoding}
L.~F.Cardoso, V.~C.Santos, R.~S.~K.Franc{\^e}s, R.~B.Prud{\^e}ncio, R.~C.Alves, Decoding machine learning benchmarks, in: Brazilian Conference on Intelligent Systems, Springer, 2020, pp. 412--425.

\bibitem{de2024explanations}
J.deSousa Ribeiro~Filho, L.~F.~F.Cardoso, R.~L.~S.daSilva, N.~J.~S.Carneiro, V.~C.~A.Santos, R.~C.deOliveira~Alves, Explanations based on item response theory (exirt): A model-specific method to explain tree-ensemble model in trust perspective, Expert Systems with Applications 244 (2024) 122986.

\bibitem{domingos2012few}
P.Domingos, A few useful things to know about machine learning, Communications of the ACM 55~(10) (2012) 78--87.

\bibitem{Dua:2019}
D.Dua, C.Graff, \href{http://archive.ics.uci.edu/ml}{{UCI} machine learning repository} (2017).
\newline\urlprefix\url{http://archive.ics.uci.edu/ml}

\bibitem{elo1978rating}
A.~E.Elo, The rating of chessplayers, past and present, Arco Pub., 1978.

\bibitem{facebook}
Facebook, \href{https://dynabench.org/about}{Rethinking ai benchmarking}.
\newline\urlprefix\url{https://dynabench.org/about}

\bibitem{ferri2009experimental}
C.Ferri, J.Hern{\'a}ndez-Orallo, R.Modroiu, An experimental comparison of performance measures for classification, Pattern Recognition Letters 30~(1) (2009) 27--38.

\bibitem{gautier2008rpy2}
L.Gautier, rpy2: A simple and efficient access to r from python, URL http://rpy. sourceforge. net/rpy2. html 3 (2008) 1.

\bibitem{glickman2012example}
M.~E.Glickman, Example of the glicko-2 system, Boston University (2012) 1--6.

\bibitem{kubat2017introduction}
M.Kubat, An introduction to machine learning, Springer, 2017.

\bibitem{lord1984comparison}
F.~M.Lord, M.~S.Wingersky, Comparison of irt true-score and equipercentile observed-score" equatings", Applied Psychological Measurement 8~(4) (1984) 453--461.

\bibitem{martinez2018dual}
F.Martinez-Plumed, J.Hernandez-Orallo, Dual indicators to analyze ai benchmarks: Difficulty, discrimination, ability, and generality, IEEE Transactions on Games 12~(2) (2018) 121--131.

\bibitem{martinez2016making}
F.Mart{\'\i}nez-Plumed, R.~B.Prud{\^e}ncio, A.Mart{\'\i}nez-Us{\'o}, J.Hern{\'a}ndez-Orallo, Making sense of item response theory in machine learning, in: Proceedings of the Twenty-second European Conference on Artificial Intelligence, 2016, pp. 1140--1148.

\bibitem{martinez2019item}
F.Mart{\'\i}nez-Plumed, R.~B.Prud{\^e}ncio, A.Mart{\'\i}nez-Us{\'o}, J.Hern{\'a}ndez-Orallo, Item response theory in ai: Analysing machine learning classifiers at the instance level, Artificial Intelligence 271 (2019) 18--42.

\bibitem{meneghetti2017application}
D.~D.~R.Meneghetti, P.~T.~A.Junior, Application and simulation of computerized adaptive tests through the package catsim, arXiv preprint arXiv:1707.03012 (2017).

\bibitem{monard2003conceitos}
M.~C.Monard, J.~A.Baranauskas, Conceitos sobre aprendizado de m{\'a}quina, Sistemas inteligentes-Fundamentos e aplica{\c{c}}{\~o}es 1~(1) (2003) 32.

\bibitem{nie2019adversarial}
Y.Nie, A.Williams, E.Dinan, M.Bansal, J.Weston, D.Kiela, Adversarial nli: A new benchmark for natural language understanding, arXiv preprint arXiv:1910.14599 (2019).

\bibitem{pedregosa2011scikit}
F.Pedregosa, G.Varoquaux, A.Gramfort, V.Michel, B.Thirion, O.Grisel, M.Blondel, P.Prettenhofer, R.Weiss, V.Dubourg, et~al., Scikit-learn: Machine learning in python, the Journal of machine Learning research 12 (2011) 2825--2830.

\bibitem{perez20152014}
D.Perez-Liebana, S.Samothrakis, J.Togelius, T.Schaul, S.~M.Lucas, A.Cou{\"e}toux, J.Lee, C.-U.Lim, T.Thompson, The 2014 general video game playing competition, IEEE Transactions on Computational Intelligence and AI in Games 8~(3) (2015) 229--243.

\bibitem{prudencio2015analysis}
R.~B.Prud{\^e}ncio, J.Hern{\'a}ndez-Orallo, A.Mart{\i}nez-Us{\'o}, Analysis of instance hardness in machine learning using item response theory, in: Second International Workshop on Learning over Multiple Contexts in ECML 2015. Porto, Portugal, 11 September 2015, Vol.~1, 2015.

\bibitem{rizopoulos2006ltm}
D.Rizopoulos, ltm: An r package for latent variable modeling and item response theory analyses, Journal of statistical software 17~(5) (2006) 1--25.

\bibitem{samothrakis2014predicting}
S.Samothrakis, D.Perez, S.~M.Lucas, P.Rohlfshagen, Predicting dominance rankings for score-based games, IEEE Transactions on Computational Intelligence and AI in Games 8~(1) (2014) 1--12.

\bibitem{smith2014reducing}
M.~R.Smith, T.Martinez, Reducing the effects of detrimental instances, in: 2014 13th International Conference on Machine Learning and Applications, IEEE, 2014, pp. 183--188.

\bibitem{song2021efficient}
H.Song, P.Flach, Efficient and robust model benchmarks with item response theory and adaptive testing., International Journal of Interactive Multimedia \& Artificial Intelligence 6~(5) (2021).

\bibitem{vanschoren2014openml}
J.Vanschoren, J.~N.Van~Rijn, B.Bischl, L.Torgo, Openml: networked science in machine learning, ACM SIGKDD Explorations Newsletter 15~(2) (2014) 49--60.

\bibitem{vevcek2014chess}
N.Ve{\v{c}}ek, M.Mernik, M.{\v{C}}repin{\v{s}}ek, A chess rating system for evolutionary algorithms: A new method for the comparison and ranking of evolutionary algorithms, Information Sciences 277 (2014) 656--679.

\end{thebibliography}

\end{document}